\documentclass[a4paper,fleqn]{cas-dc}
\usepackage{xurl} % use {xurl} instead of {url} for auto line break
\usepackage{color}
\usepackage{caption}
\usepackage{subcaption}
\usepackage{amsmath}
\usepackage{makecell}
\usepackage{multirow}
\usepackage{tikz}
\usepackage[authoryear,longnamesfirst]{natbib}

\def\tsc#1{\csdef{#1}{\textsc{\lowercase{#1}}\xspace}}
\tsc{WGM}
\tsc{QE}
\tsc{EP}
\tsc{PMS}
\tsc{BEC}
\tsc{DE}
\newcommand{\subsec}[1]{\noindent{\textbf{#1~~}}}
\newcommand{\ie}{i.e.\xspace}
\newcommand{\eg}{e.g.\xspace}
\newcommand{\class}[1]{\ensuremath{\mathsf{#1}}}
\usepackage{mathtools}       % For math equations

\DeclareMathOperator*{\argmin}{arg\,min}

\newcommand{\x}{\mathbf{x}}

\newcommand{\gD}{\mathcal{D}}
\newcommand{\gP}{\mathcal{P}}

\newcommand{\layer}[1]{\ensuremath{\mathsf{#1}\xspace}}

\newcommand{\layerunit}[2]{\ensuremath{\mathsf{#1_{#2}}\xspace}}

\DeclareRobustCommand\mycircle[1]{\tikz\draw[black,fill=#1] (0,0) circle (.6ex);}
\DeclareRobustCommand\mytriangle[1]{\tikz{\filldraw[black,fill=#1] (0,0) --
(0.2cm,0) -- (0.1cm,0.2cm);}}
\DeclareRobustCommand\mysquare[1]{\tikz\draw[black,fill=#1] (0,0) rectangle ++(6pt,6pt);}

\begin{document}
\let\WriteBookmarks\relax
\def\floatpagepagefraction{1}
\def\textpagefraction{.001}

\shorttitle{Biases in Adversarially Robust CNNs}

\shortauthors{Chen et~al.}
% \journal{Vision Research}
\title [mode = title]{The shape and simplicity biases of adversarially robust ImageNet-trained CNNs}                      
% \tnotemark[1,2]

% \tnotetext[1]{Fund can be added here.}

% \tnotetext[2]{The second title footnote which is a longer text matter
%   to fill through the whole text width and overflow into
%   another line in the footnotes area of the first page.}

% \author[1,3]{CV Radhakrishnan}[type=editor,
%                         auid=000,bioid=1,
%                         prefix=Sir,
%                         role=Researcher,
%                         orcid=0000-0001-7511-2910]
\author{Peijie Chen}[
% type=editor,
                        % auid=000,bioid=1,
                        % orcid=0000-0002-0621-4689
                        ]
% \cormark[1]

% % Footnote of the first author

\ead{peijie.chen.auburn@gmail.com}

% \ead[url]{www.cvr.cc, cvr@sayahna.org}

\credit{Conceptualization of this study, Methodology, Software}

\author[]{Chirag Agarwal}

\ead{chirag.agarwall12@gmail.com}
\author[]{Anh Nguyen}[
type=editor,
%
% orcid=0000-0003-0528-9416
%   role=Co-ordinator,
%   suffix=Jr,
   ]
\cormark[1]
\ead{anh.ng8@gmail.com}
% \ead[URL]{Auburn Unive}

% \affiliation{organization={Auburn University},
%     addressline={Department of Computer Science and Software Engineering}, 
%     city={Auburn},
%     % citysep={}, % Uncomment if no comma needed between city and postcode
%     postcode={36849}, 
%     state={AL},
%     country={USA}}

\credit{Data curation, Writing - Original draft preparation}

% \affiliation[2]{organization={Sayahna Foundation},
%     % addressline={}, 
%     city={Jagathy},
%     % citysep={}, % Uncomment if no comma needed between city and postcode
%     postcode={695014}, 
%     state={Trivandrum},
%     country={India}}

% \author%
% [1,3]
% {Rishi T.}
% \cormark[2]
% \fnmark[1,3]
% \ead{rishi@stmdocs.in}
% \ead[URL]{www.stmdocs.in}

% \affiliation[3]{organization={STM Document Engineering Pvt Ltd.},
%     addressline={Mepukada}, 
%     city={Malayinkil},
%     % citysep={}, % Uncomment if no comma needed between city and postcode
%     postcode={695571}, 
%     state={Trivandrum},
%     country={India}}

\cortext[cor1]{Corresponding author. \\All authors from \textbf{Auburn University}.}
% \cortext[cor2]{Principal corresponding author}

% \fntext[fn1]{This is the first author footnote. but is common to third
% \fntext[fn2]{Another author footnote, this is a very long footnote and
%   it should be a really long footnote. But this footnote is not yet
%   sufficiently long enough to make two lines of footnote text.}

% % For a title note without a number/mark
% \nonumnote{This note has no numbers. In this work we demonstrate $a_b$
%   the formation Y\_1 of a new type of polariton on the interface
%   between a cuprous oxide slab and a polystyrene micro-sphere placed

\begin{abstract}
% This template helps you to create a properly formatted \LaTeX\ manuscript.
% \noindent\texttt{\textbackslash begin{abstract}} \dots 
% \texttt{\textbackslash end{abstract}} and
% \verb+\begin{keyword}+ \verb+...+ \verb+\end{keyword}+ 
% which
% contain the abstract and keywords respectively. 
% \noindent Each keyword shall be separated by a \verb+\sep+ command.
Increasingly more similarities between human vision and convolutional neural networks (CNNs) have been revealed in the past few years.
Yet, vanilla CNNs often fall short in generalizing to adversarial or out-of-distribution (OOD) examples which humans demonstrate superior performance.
Adversarial training is a leading learning algorithm for improving the robustness of CNNs on adversarial and OOD data; however, little is known about the properties, specifically the shape bias and internal features learned inside adversarially-robust CNNs.
% However, little is known adversarially-robust CNNs, a class of CNNs that demonstrate state-of-the-art performance against adversarial and  (OOD) examples.
% Adversarial training has been the topic of dozens of studies and a leading method for defending against adversarial attacks.
% Yet, it remains largely unknown (a) how adversarially-robust ImageNet classifiers (R classifiers) generalize to out-of-distribution examples; and (b) how their generalization capability relates to their hidden representations.
In this paper, we perform a thorough, systematic study to understand the shape bias and some internal mechanisms that enable the generalizability of AlexNet, GoogLeNet, and ResNet-50 models trained via adversarial training.
We find that while standard ImageNet classifiers have a strong texture bias, their R counterparts rely heavily on shapes.
Remarkably, adversarial training induces three simplicity biases into hidden neurons in the process of ``robustifying'' CNNs.
That is, each convolutional neuron in R networks often changes to detecting (1) pixel-wise smoother patterns \ie a mechanism that blocks high-frequency noise from passing through the network; (2) more lower-level features \ie textures and colors (instead of objects);and (3) fewer types of inputs. 
Our findings reveal the interesting mechanisms that made networks more adversarially robust and also explain some recent findings \eg why R networks benefit from much larger capacity \citep{xie2020intriguing} and can act as a strong image prior in image synthesis \citep{santurkar2019image}. 

\end{abstract}

% \includegraphics{figs/grabs.pdf}

% Research highlights0
% \begin{highlights}
% \item Adversarially-robust CNNs prefer shapes while traditional CNNs prefer textures
% \item Adversarially-robust CNNs outperform traditionial CNNs on some types of distorted images (including texture-less silhouettes)
% \item Adversarial training cause neurons to be more simple and less multifaceted.
% \end{highlights}

% Keywords
\begin{keywords}
Deep Learning \sep Computer Vision \sep Robustness \sep Adversarial Training \sep Interpretability \sep Texture vs. Shape \sep
\end{keywords}
\maketitle
\section{Introduction}
Convolutional neural networks (CNNs), specifically, AlexNet \citep{krizhevsky2012imagenet}, have been increasingly found to exhibit interesting correspondences with human object recognition \citep{rajalingham2018large,serre2019deep,cichy2016comparison}.
However, a remarkable difference between ImageNet-trained CNNs and humans is that while the CNNs leverage mostly texture cues to categorize images, humans recognize an object by relying on its outline or silhouette
\citep{geirhos2018imagenet}.
Interestingly, this shape bias of human vision was hypothesized to enable the superior performance of humans on out-of-distribution (OOD) data \citep{geirhos2018generalisation}, which CNNs often misclassify.

For example, simply adding Gaussian noise to the input image can dramatically reduces AlexNet's top-1 accuracy from $\sim$57\% to $\sim$11\% \citep{hendrycks2018benchmarking}.
More severely, adding imperceptible, pixel-wise changes to an input image of a school bus yields a visually-identical image that would cause state-of-the-art CNNs to mislabel ``ostrich'' \citep{szegedy2013intriguing-properties-of-neural}.
An interesting class of CNNs that are state-of-the-art on such OOD noisy or adversarial data are adversarially-robust CNNs (hereafter, R networks), \ie the networks that are trained to correctly label adversarial examples instead of real examples \citep{madry2017towards}.
That is, adversarial training has substantially improved model robustness to some types of out-of-distribution and adversarial data \citep{madry2017towards, xie2020adversarial}.

Therefore, we aim to understand the relationship between the OOD performance and shape bias of R networks by answering three main questions.
First, it remains unknown whether a \emph{shape} bias exists in a R network trained on the large-scale ImageNet \citep{russakovsky2015imagenet}, which often induces a large \emph{texture} bias into vanilla CNNs \citep{geirhos2018imagenet}, \eg to separate $\sim$150 four-legged species in ImageNet. 
Thus, an important question is:

\emph{\textbf{Q1:} On ImageNet, do adversarially-robust networks (a.k.a., R networks) prefer shapes over textures?}

% This shape bias was hypothesized to enable the superior performance of humans on out-of-distribution (OOD) data \citep{geirhos2018generalisation}, compared to state-of-the-art convolutional neural networks (CNNs), which rely heavily on textures \citep{geirhos2018imagenet}.
% Interestingly, the recently reported similarities between AlexNet \cite{krizhevsky2012imagenet} and human vision 

% A current, large gap between modern machine and human intelligence is in flexible, systematic, OOD generalization \citep{goyal2020inductive}.
% For example, despite excellent test-set performance, CNNs often misclassify OOD examples \citep{nguyen2015deep}
% including \emph{``adversarial examples''}, \ie modified inputs that are imperceptibly different from the real data but change predicted labels entirely \citep{szegedy2013intriguing-properties-of-neural}.

% That is, adversarially-robust CNNs substantially outperform vanilla CNNs 
% ---has been a leading method 
% \citep{madry2017towards}
% \citep{goodfellow-2014-arXiv-explaining-and-harnessing-adversarial,madry2017towards} 
% in defending against adversarial attacks \citep{athalye2018obfuscated}.
% Interestingly, test-set accuracy can be improved when real images are properly incorporated into adversarial training \citep{xie2020adversarial}.

\noindent This question is interestingly also because \cite{santurkar2019image} found that ImageNet-trained R networks act as a strong image \emph{texture} prior, \ie they can be successfully used for many pixel-wise image optimization tasks .

% Also, this shape-bias hypothesis suggested by \cite{zhang2019interpreting} from findings on smaller-datasets seems to contradict the recent results that R networks on ImageNet act as a strong \emph{texture} prior \ie they can be successfully used for many image translation tasks without any extra image prior \citep{santurkar2019image}.
% The above discussion leads to a follow-up question:

% This important link was not yet established.
Second, \citet{geirhos2018imagenet} found that CNNs trained to be more \emph{shape}-biased (but \emph{not} via adversarial training) can generalize better to many unseen ImageNet-C \citep{hendrycks2018benchmarking}
image corruptions than S networks, which have a much weaker shape bias \citep{brendel2018bagnets}.
Therefore, we ask:

\emph{\textbf{Q2:} If an R network has a stronger preference for shapes than standard ImageNet-trained CNNs (hereafter, S networks), will it perform better on OOD images?}

% In contrast, there was also evidence that classifiers trained on one type of images often do not generalize well to others \citep{geirhos2018generalisation,nguyen2015deep}.
\noindent It is worth noting that R networks, which were trained solely on adversarial examples, often \emph{underperform} S networks on real test sets \citep{tsipras2018robustness}. 
Beside being more robust to adversarial attacks, some CNNs trained via AdvProp \citep{xie2020adversarial} (a variant of \cite{madry2017towards}'s adversarial training framework that trains CNNs on both real and adversarial data) obtain a similarly high accuracy on real data as S networks; however, it is unknown whether AdvProp CNNs are shape- or texture-biased (which we answer in Table~\ref{tab:shp_vs_tex}).

% perhaps due to an inherent trade-off \citep{madry2017towards}, a mismatch between real vs. adversarial distributions \citep{xie2020adversarial}, or a limitation in architectures---AdvProp did not improve the test-set performance of ResNets \citep{xie2020adversarial}.

Third, while most prior work studied the behaviors of R CNNs as a black-box classifier, little is known about the internal network characteristics of R networks and, what enable their shape bias and OOD performance. 
Therefore, we attempt to shed light into the last question:

\emph{\textbf{Q3:} How did adversarial training change the hidden neural representations to make CNNs more shape-biased and adversarially-robust?}

In this paper, we harness two common datasets in ML interpretability and neuroscience---cue-conflict \citep{geirhos2018imagenet} and NetDissect \citep{bau2017network}---to answer the three questions above via a systematic study across three different convolutional architectures---AlexNet \citep{krizhevsky2012imagenet}, GoogLeNet \citep{szegedy2015going}, and ResNet-50 \citep{he2016deep}---trained to perform image classification on the large-scale ImageNet dataset \citep{russakovsky2015imagenet}.
% Our study reveals some interesting connections between the generalization capability of robust networks and their inner functions.
We find that:\footnote{All code and data will be available on github upon publication.}

\begin{enumerate}
    \item R classifiers trained on ImageNet prefer shapes over textures $\sim$67\% of the time---a stark contrast to S classifiers, which prefer shapes at $\sim$25\% (Sec.~\ref{sec:shapes_or_textures}).
    % \item While R classifiers largely underperform S classifiers on shape-less distorted images (patch-shuffled images), they slightly, but consistently, outperform standard counterparts on texture-less distorted images (stylized and silhouetted images) (Sec.~\ref{sec:texture-less}).
    \item Consistent with the strong shape bias, R classifiers interestingly outperform S counterparts on texture-less, distorted images (stylized and silhouetted images) (Sec.~\ref{sec:texture-less}).
    \item Adversarial training makes CNNs more robust by (a) blocking pixel-wise input noise via smooth filters (Sec.~\ref{sec:smooth_tv}); (b) reducing the set of high-activating inputs to simple patterns (Sec.~\ref{sec:netdissect_robust_difference}).
    % \item Given identical architectures, $\sim$10\% to 15\% of the neurons in a R network shift roles to detecting lower-level features \ie colors and textures instead of objects (Sec.~\ref{sec:netdissect_robust_difference}).
    % \item Each R unit is often simpler \ie detecting less unique features than standard units (Sec.~\ref{sec:simpler_units}).
    \item Units that detect texture patterns (according to NetDissect) are not only useful to texture-based image classification but can be also highly useful to \emph{shape}-based classification (Sec.~\ref{sec:R_net_shift_to_shape}). 
    \item By aligning NetDissect and cue-conflict frameworks, we found that hidden neurons in R networks are surprisingly neither strongly shape-biased nor texture-biased, but instead generalists that detect low-level features (Sec.~\ref{sec:R_net_shift_to_shape}).
\end{enumerate}

The exciting differences between human vision and R CNNs revealed by our work can inform future research into what enables the strong generalization capability of human vision and how to improve state-of-the-art CNNs.

\section{Networks and Datasets}
\label{sec:networks_datasets}

\subsec{Networks}
To understand the effects of adversarial training across a wide range of architectures, we compare each pair of S and R models while keeping their network architectures constant.
That is, we conduct all experiments on two groups of classifiers: (a) standard AlexNet, GoogLeNet, \& ResNet-50 (hereafter, ResNet) models pre-trained on the 1000-class 2012 ImageNet dataset; and (b) three adversarially-robust counterparts \ie AlexNet-R, GoogLeNet-R, \& ResNet-R which were trained via adversarial training (see below).

\subsec{Training} 
A standard classifier with parameters  $\theta$ was trained to minimize the cross-entropy loss $L$ over pairs of (training example $x$, ground-truth label $y$) drawn from the ImageNet training set $\gD$:

% \vspace{-0.3cm}
\begin{equation}
    \label{eq:vanilla_training}
    \argmin_{\theta}
    \mathbb{E}_{(x, y) \sim \gD}
    \Bigl[ 
    L(\theta, x, y)
    \Bigr]
\end{equation}

On the other hand, we trained each R classifier via \citet{madry2017towards} adversarial training framework where each real example $x$ is changed by a perturbation $\Delta$:

% \vspace{-0.2cm}
\begin{equation}
\label{eq:adv_training}
\argmin_{\theta}
\mathbb{E}_{(x, y) \sim \gD}
\Bigl[ 
\max_{\Delta \in \gP}
L(\theta, x + \Delta, y)
\Bigr]
\end{equation}

\noindent and $\gP$ is the perturbation range allowed within an $L_2$ norm.

\subsec{Hyperparameters} The S models were downloaded from PyTorch model zoo \citep{torchvis88:online}. 
We adversarially trained all R models using the \layer{robustness} library \citep{robustness}, using \emph{the same} hyperparameters in \citep{engstrom2020adversarial,santurkar2019image,bansal2020sam} because these previous works have shown that adversarial training significantly changed the inner-workings of ImageNet CNNs---\ie becoming a strong image prior with perceptually-aligned deep features.
That is, adversarial examples were generated using Projected Gradient Descent (PGD) \citep{madry2017towards} with an $L_2$ norm constraint $\epsilon$ of $3$, a step size of 0.5,
and 7 PGD-attack steps. 
R models were trained using an SGD optimizer for 90 epochs with a momentum of $0.9$, an initial learning rate of $0.1$ (which is reduced 10 times every $30$ epochs), a weight decay of $10^{-4}$, and a batch size of 256 on 4 Tesla-V100 GPU’s.

% IJCV version table 1
% \vspace{-0.3cm}
\begin{table*}[t]
   \caption{
   Top-1 accuracy (\%) on 50K-image ImageNet validation-set and PGD adversarial images (number of iterations = 7; $\epsilon$ = 3; step size = 0.5).
%  \todo{Update this table with new numbers!}
%  \peijie{Table updated}
   }
   \label{tab:val_adv_acc}
   \def\arraystretch{1.0}%
   \begin{center}
      \begin{tabular}{l|c|c|c|c|c|c|c|c}
%        \hline
%        & AlexNet & AlexNet-R & GoogLeNet & GoogLeNet-R & ResNet & ResNet-R & AdvProp PGD1 & AdvProp PGD5\\
%        \hline
         \hline
          Architecture& \multicolumn{2}{c|}{AlexNet} &  \multicolumn{2}{c|}{GoogLeNet} &  \multicolumn{4}{c}{ResNet-50} \\ \hline
          Training algorithm& Standard & Robust & Standard & Robust & Standard & AdvProp PGD1 & AdvProp PGD5 & Robust \\ \hline
%        ImageNet(old) & \textbf{56.52} & 39.83 & \textbf{68.86} & 50.94 & \textbf{75.59} & 56.25 \\ \hline
         ImageNet & 56.52 & 39.83 & 69.78 & 43.57 & 76.13 & 77.31 & 77.01 & 57.90 \\ \hline        
         Adversarial & 0.18 & 22.27 & 0.08 & 31.23 & 0.35 & 69.02 & 73.55 & 36.11 \\ \hline
      \end{tabular}
   \end{center}
\end{table*}

\begin{table*}[ht]
   \caption{
   While Standard classifiers rely heavily on textures, Robust classifiers rely heavily on shapes. 
   Texture and Shape bias scores are the top-1 accuracy (\%) computed on the cue-conflict dataset proposed by \citet{geirhos2018imagenet}.
   }
   \label{tab:shp_vs_tex}
   \def\arraystretch{1.1}%
   \begin{center}
      \begin{tabular}{l|c|c|c|c|c|c|c|c}
         \hline
          Architecture& \multicolumn{2}{c|}{AlexNet} &  \multicolumn{2}{c|}{GoogLeNet} &  \multicolumn{4}{c}{ResNet-50} \\ \hline
          Training algorithm & Standard & Robust & Standard & Robust & Standard & AdvProp PGD1 & AdvProp PGD5 & Robust \\ \hline
          Training data & Real & Adv. & Real & Adv. & Real & Real + Adv. & Real + Adv. & Adv. \\ \hline
         Texture & \textbf{73.61} & 34.67 & \textbf{74.91} & 34.43 & \textbf{77.79} & \textbf{68.24}  & \textbf{63.11} & 29.63 \\ \hline
         Shape & 26.39 & \textbf{65.32} & 25.08 & \textbf{65.56} & 22.20 & 31.75 & 36.89 & \textbf{70.36} \\ \hline
      \end{tabular}
   \end{center}
\end{table*}

% We obtained the two regular models from the PyTorch model zoo \citep{torchvis88:online}, the ResNet-R from \citep{engstrom2019adversarial}, and we trained GoogLeNet-R by ourselves using the code released by \citep{engstrom2019adversarial}.

Compared to the standard counterparts, R models have substantially higher adversarial accuracy but lower ImageNet validation-set accuracy (Table~\ref{tab:val_adv_acc}).
To compute adversarial accuracy, we perturbed validation-set images with the same PGD attack settings as used in training.

% \anh{Chirag, can you add a Table that reports both original ImageNet validation + adversarial performance for all 6 networks?}\chirag{Done}
% ICLR version table 1
% \vspace{-0.3cm}
% \begin{table}[ht]
%  \caption{
%  Top-1 accuracy (\%) on 50K-image ImageNet validation-set and PGD adversarial examples.
% %   \todo{Update this table with new numbers!}
% %   \peijie{Table updated}
%  }
%  \label{tab:val_adv_acc}
%  \def\arraystretch{1.1}%
%  \begin{center}
%     \begin{tabular}{l|c|c|c|c|c|c}
%        \hline
%        & AlexNet & AlexNet-R & GoogLeNet & GoogLeNet-R & ResNet & ResNet-R \\
%        \hline
% %         ImageNet(old) & \textbf{56.52} & 39.83 & \textbf{68.86} & 50.94 & \textbf{75.59} & 56.25 \\ \hline
%        ImageNet & \textbf{56.52} & 39.83 & \textbf{69.78} & 43.57 & \textbf{76.13} & 57.90 \\ \hline         
%        Adversarial & 0.18 & \textbf{22.27} & 0.08 & \textbf{31.23} & 0.35 & \textbf{36.11} \\ \hline
%     \end{tabular}
%  \end{center}
% \end{table}

\subsec{Correctly-labeled image subsets: ImageNet-CL}
Following \cite{bansal2020sam}, to compare the behaviors of two networks of identical architectures on the same inputs, we tested them on the largest ImageNet validation subset (hereafter, ImageNet-CL) where both models have 100\% accuracy.
The sizes of the three subsets for three architectures---AlexNet, GoogLeNet, and ResNet---are respectively: 17,693, 20,238, and 27,343.
% Peijie: ImageNet-CL numbers replaced with new ImageNet-CL (Using new evaluation script)
% 17,693, 24,581, and 27,343.
On modified ImageNet images (\eg, adversarial in Fig.~\ref{fig:unseen}b), we only tested each CNN pair on the modified images whose original versions exist in ImageNet-CL.
That is, we wish to gain deeper insights into how CNNs behave on correctly-classified images, and then how their behaviors change when some input feature (\eg textures or shapes) is modified.

% To study the effects of adversarial training, we hold network architectures constant.
% Because adversarial training might require a much larger model capacity \citep{xie2020intriguing}, 
% To understand how a network behavior changes when some information (\eg textures or shapes) in a given ImageNet image is changed, in all experiments with modified images, we only tested each pair of networks on the modified images whose original ImageNet versions were correctly labeled by both networks.
% That is, following \cite{bansal2020sam,geirhos2018imagenet}, for each architecture, we used an ImageNet validation subset where both models have 100\% accuracy.
% The sizes of the three subsets (hereafter, ImageNet-CL) for three architectures---AlexNet, GoogLeNet, and ResNet---are respectively: 17,693, 24,581, and 27,343.

\section{Experiment and Results}
% \chirag{I feel the subsections do not come up as subsections essentially..maybe use textbf?}
% \peijie{I feel the same way. But since we cannot change the IJCV format so I will just leave it..}
\subsection{Do ImageNet adversarially-robust networks prefer shapes or textures?}
\label{sec:shapes_or_textures}

It is important to know which input feature a classifier uses when making decisions.
While standard ImageNet networks often carry a strong texture bias \citep{geirhos2018imagenet}, it is unknown whether their adversarially-robust counterparts would be heavily texture- or shape-biased.
% We hypothesized that adding pixel-wise adversarial noise to training images might encourage R networks to rely more on global features (\ie shapes) to make predictions.
Here, we test this hypothesis by comparing S and R models on the well-known cue-conflict dataset \citep{geirhos2018imagenet}.
That is, we feed ``stylized'' images provided by \citet{geirhos2018imagenet} that contain contradicting texture and shape cues (\eg elephant skin on a cat silhouette) and count the times a model uses textures or shapes (\ie outputting \class{elephant} or \class{cat}) when it makes a correct prediction.

\subsec{Experiment}
Our procedure follows \cite{geirhos2018imagenet}.
First, we excluded 80 images that do not have conflicting cues (e.g. cat textures on cat shapes) from their 1,280-image dataset.
Each texture or shape cue belongs to one of 16 MS COCO \citep{caesar2018coco} coarse labels (\eg \class{cat} or \class{elephant}).
Second, we ran the networks on these images and converted their 1000-class probability vector outputs into 16-class probability vectors by taking the average over the probabilities of the fine-grained classes that are under the same COCO label.
Third, we took only the images that each network correctly labels (\ie into the texture or shape class), which ranges from 669 to 877 images (out of 1,200) for 6 networks and computed the texture and shape accuracies over 16 classes.

% We tested each model on the entire cue-conflict dataset of 1,280 stylized images \citep{geirhos2018imagenet}.
% \anh{I think we computed the accuracy scores below on the set of cue-conflict images that both models correctly classify, right?}

% \chirag{The cue-conflict dataset was presented in \citep{geirhos2018imagenet} where the images were generated using the iterative style transfer \citep{gatys2016image} between an image of the texture dataset (as style) and an image from the original data set (as content). They created a total of 1280 cue conflict images.}

\subsec{Results} 
On average, over three architectures, R classifiers rely on shapes $\geq$ 67.08\% of the time, \ie $\sim$2.7$\times$ higher than 24.56\% of the S models (Table~\ref{tab:shp_vs_tex}).
In other words, by replacing real with adversarial examples, adversarial training causes the heavy texture bias of ImageNet classifiers \citep{geirhos2018imagenet,brendel2018bagnets} to drop substantially ($\sim$2.7$\times$).
%% S resnet50 on ImageNet top1: 76.130%
%% AdvProp pgd 1 on ImageNet top1: 77.314%
%% AdvProp pgd 5 on ImageNet top1: 77.012%
Additionally, we also test two adversarially-robust models that are trained on a mix of real and adversarial data via AdvProp \citep{xie2020adversarial} with a PGD $L_2$ $\epsilon$ of 1 and 5, respectively (see full hyperparameters in the \href{https://github.com/tingxueronghua/pytorch-classification-advprop}{training code}).
These AdvProp models obtain a similarly high accuracy on both validation as well as adversarial images (Table~\ref{tab:val_adv_acc}; PGD1 \& PGD5).
Interestingly, AdvProp models are also heavily texture-biased (Table~\ref{tab:shp_vs_tex}; 68.24\% vs. 31.75\%) but their preferences are in between that of the vanilla and R models.
Moreover, as CNNs are trained with increasingly more adversarial perturbations and less real data (from Standard $\to$ PGD1 $\to$ PGD5 $\to$ Robust), the texture bias also decreases (from 77.79\%$\to$ 68.24\% $\to$ 63.11\% $\to$ 29.63\%; Table~\ref{tab:shp_vs_tex}).
This result presents a strong evidence that \textbf{real ImageNet images strongly induce a texture bias} and \textbf{adversarial images induce a shape bias} into CNNs.

\begin{table*}[t]
\caption{
    Adversarially-robust (R) models outperform vanilla (S) models when both are under PGD-adversarial attacks (b). R models consistently outperform S model in texture-less (e,f) distortions.
    % While R models do not generalize well to common distortions (c), yet, they interestingly outperform S models on texture-less images (d--f).
   Here, we report top-1 accuracy scores (\%) on the transformed images whose original real versions were correctly-labeled (a) by both S and R models.
%  ``ImageNet-C'' column (c) shows the mean accuracy scores over all 15 distortion types (see breakdown accuracy on each distortion type in Table~\ref{tab:imagenet_c}). 
   ``Scrambled'' column (c) shows the mean accuracy scores over three patch-scrambling types (details in Fig.~\ref{fig:accuracy_scrambled_images}).
}
\label{tab:adv_acc}
\centering
\begin{tabular}{l|c|c|c|c|c|cc}
\hline
\multirow{2}{*}{Network} & \multirow{2}{*}{ \makecell{(a)\\ Real \\ ImageNet} } & \multirow{2}{*}{ \makecell{(b)\\ Adversarial}} & Shape-less & \multicolumn{3}{c}{Texture-less} \\ \cline{4-7}
 & &  & \makecell{(c) \\ Scrambled} & \makecell{(d) \\ Stylized} & \makecell{(e) \\ B\&W} & \makecell{(f) \\ Silhouette} \\\hline
 AlexNet & 100 & 0.18 & \textbf{34.59} & 6.31 & 20.08 & 7.72\\
AlexNet-R & 100 & \textbf{22.27} & 16.92 & \textbf{9.11} & \textbf{35.25} & \textbf{9.30}\\ \hline
GoogLeNet & 100 & 0.08 & \textbf{49.74} & 13.74 & 43.48 & 10.17\\
GoogLeNet-R & 100 & \textbf{31.23} & 31.15 & 12.54 & \textbf{44.55} & \textbf{24.12}\\ \hline
ResNet & 100 & 0.35 & \textbf{58.04} & 10.68 & 16.96 & 3.95\\
ResNet-R & 100 & \textbf{36.11} & 34.46 & \textbf{15.62} & \textbf{53.89} & \textbf{22.30} \\
% \multirow{2}{*}{Network} & \multirow{2}{*}{ \makecell{(a)\\ Real \\ ImageNet} } & \multirow{2}{*}{ \makecell{(b)\\ Adversarial}} & & Shape-less & \multicolumn{2}{c}{Texture-less} \\ \cline{4-8}
%  & &  & \makecell{(c) \\ ImageNet-C} & \makecell{(d) \\ Scrambled} & \makecell{(e) \\ Stylized} & \makecell{(f) \\ B\&W} & \makecell{(g) \\ Silhouette} \\\hline
% AlexNet & 100 & 0.18 & 35.06 & \textbf{34.59} & 6.31 & 20.08 & 7.72\\
% AlexNet-R & 100 & \textbf{22.27} & \textbf{44.94} & 16.92 & \textbf{9.11} & \textbf{35.25} & \textbf{9.30}\\ \hline
% GoogLeNet & 100 & 0.08 & \textbf{58.27} & \textbf{49.74} & 13.74 & 43.48 & 10.17\\
% GoogLeNet-R & 100 & \textbf{31.23} & 38.76 & 31.15 & 12.54 & \textbf{44.55} & \textbf{24.12}\\ \hline
% ResNet & 100 & 0.35 & 51.70 & \textbf{58.04} & 10.68 & 16.96 & 3.95\\
% ResNet-R & 100 & \textbf{36.11} & 51.72 & 34.46 & \textbf{15.62} & \textbf{53.89} & \textbf{22.30} \\
\hline
\end{tabular}
\end{table*}

\begin{figure*}[t]
   \centering
    % AlexNet: 
    % Real: GT: red-backed sandpiper ; prob: 1.000 | Predicted red-backed sandpiper ; prob: 1.000
    % Adv:  GT: red-backed sandpiper ; prob: 0.000 | Predicted redshank ; prob: 0.972
    % imagenet_c:  GT: red-backed sandpiper ; prob: 0.633 | Predicted red-backed sandpiper ; prob: 0.633
    % scrambled:  GT: red-backed sandpiper ; prob: 0.715 | Predicted red-backed sandpiper ; prob: 0.715
    % stylized: GT: red-backed sandpiper ; prob: 0.007 | Predicted bonnet ; prob: 0.200
    % silhouette :  GT: red-backed sandpiper ; prob: 0.000 | Predicted American egret ; prob: 0.453
    
    % AlexNet-R:
    % Real: GT: red-backed sandpiper ; prob: 0.830 | Predicted red-backed sandpiper ; prob: 0.830
    % Adv: GT: red-backed sandpiper ; prob: 0.817 | Predicted red-backed sandpiper ; prob: 0.817
    % imagenet_c:  GT: red-backed sandpiper ; prob: 0.699 | Predicted red-backed sandpiper ; prob: 0.699
    % scrambled:  GT: red-backed sandpiper ; prob: 0.237 | Predicted red-backed sandpiper ; prob: 0.237
    % stylized:  GT: red-backed sandpiper ; prob: 0.297 | Predicted red-backed sandpiper ; prob: 0.297
    % silhouette :   GT: red-backed sandpiper ; prob: 0.000 | Predicted American egret ; prob: 0.238
   {
%     \begin{flushleft}
%         \small
%        \hspace{0.5cm}(a) Real 
%        \hspace{0.8cm}(b) Adversarial
%        \hspace{0.4cm}(c) ImageNet-C
%        \hspace{0.4cm}(d) Scrambled
%        \hspace{0.4cm}(e) Stylized
%        \hspace{0.8cm}(f) Silhouette
%     \end{flushleft}
      \begin{flushleft}
          \small
         \hspace{1cm}(a) Real 
         \hspace{1.2cm}(b) Adversarial
         \hspace{0.8cm}(c) Scrambled
         \hspace{0.9cm}(d) Stylized
         \hspace{1.1cm}(e) B\&W
         \hspace{1.1cm}(f) Silhouette
%        \hspace{1cm}(a) Real 
%        \hspace{0.6cm}(b) Adversarial
%        \hspace{0.2cm}(c) ImageNet-C
%        \hspace{0.2cm}(d) Scrambled
%        \hspace{0.45cm}(e) Stylized
%        \hspace{0.77cm}(f) B\&W
%        \hspace{0.8cm}(g) Silhouette
      \end{flushleft}
   }
%  \vspace{-0.3cm}
%  {
%  \color{blue}
%     \begin{flushleft}
%         \small
%        \hspace{0.1cm} \class{sandpiper~1.0}
%        \hspace{0.25cm} \class{redshank~0.972}
%        \hspace{0.2cm} \class{sandpiper~0.633}
%        \hspace{0.1cm} \class{sandpiper~0.715}
%        \hspace{0.1cm} \class{bonnet~0.200}
%        \hspace{0.6cm} \class{egret~0.453}
%     \end{flushleft}
%  }
   \includegraphics[width=0.95\linewidth]{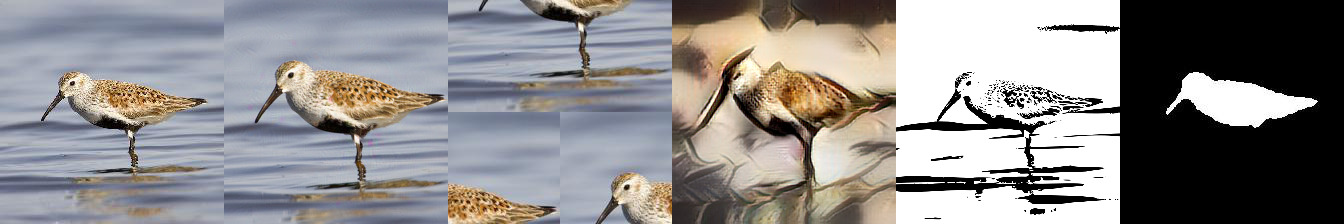}
%  \vspace{-0.4cm}
%  {
%  \color{orange}
%     \begin{flushleft}
%         \small
%        \hspace{0.01cm} \class{sandpiper~0.830}
%        \hspace{0.1cm} \class{sandpiper~0.817}
%        \hspace{0.1cm} \class{sandpiper~0.699}
%        \hspace{0.1cm} \class{sandpiper~0.237}
%        \hspace{0.1cm} \class{sandpiper~0.297}
%        \hspace{0.3cm} \class{egret~0.238}
%     \end{flushleft}
%  }  
   \caption{
      Example distorted images (b--g).
    %    We show an example Gaussian-noise-added image (c) out of all 15 ImageNet-C types.
      See Fig.~\ref{fig:shape_texture-less} for more examples.
      }
   \label{fig:unseen}
\end{figure*}

\subsection{Do robust networks generalize to unseen types of distorted images?}
\label{sec:distorted}

\cite{geirhos2018imagenet} found that some training regimes that encourage CNNs to focus more on \emph{shape} can improve their performance on unseen image distortions.
Sec.~\ref{sec:shapes_or_textures} shows that training with more adversarial examples makes a CNN more shape-biased.
Therefore, it is interesting here to understand how adversarial training improves CNN generalization performance.

That is, we test R models on two types of \emph{controlled} images where either shape or texture cues are removed from the original, correctly-labeled ImageNet images (Fig.~\ref{fig:unseen}c--f).
Note that when both shape and texture cues are present, \eg in cue-conflict images, R classifiers consistently prefer shape over texture, \ie a shape \emph{bias} (Table~\ref{tab:shp_vs_tex}).
However, this bias may not necessarily carry over to the images where only either texture or shape cues are present.

\subsubsection{Performance on shape-less images}
\label{sec:distorted_shape}

% Our previous experiments consistently suggest a strong shape preference in R models.

\begin{figure*}[t]
   \centering
      \begin{subfigure}{0.49\linewidth}
        % \begin{subfigure}
      % \vspace*{-0.6cm}
      \centering
      {
      \begin{flushleft}
         \hspace{-0.1cm}\rotatebox{90}{\hspace{-6.5cm}ResNet-R\hspace{2.5cm}{ResNet}}
         \hspace{0.7cm}$1\times1$
         \hspace{1.2cm}$2\times2$
         \hspace{1.2cm}$4\times4$
         \hspace{1.2cm}$8\times8$
      \end{flushleft}
       }
      \includegraphics[width=0.95\linewidth]{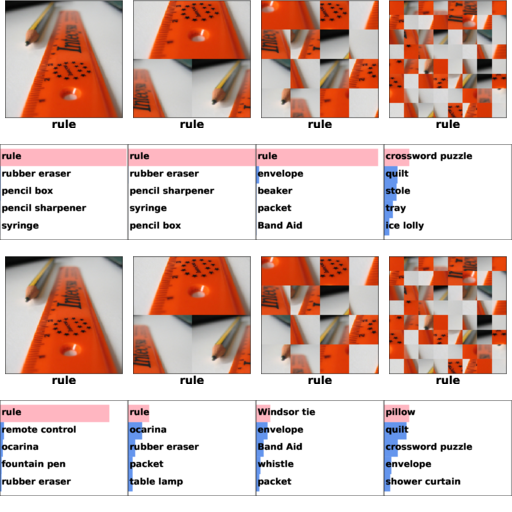}
%     \caption{Top-5 predictions on the scrambled versions of a \class{ruler} image. 
%     }
%     \label{fig:resnet_scrambled_ruler}
   \end{subfigure}
   \centering
      \begin{subfigure}{0.49\linewidth}
      % \vspace*{-0.6cm}
      \centering
      {
      \begin{flushleft}
         \hspace{-0.1cm}\rotatebox{90}{\hspace{-6.5cm}ResNet-R\hspace{2.5cm}{ResNet}}
         \hspace{0.7cm}$1\times1$
         \hspace{1.2cm}$2\times2$
         \hspace{1.2cm}$4\times4$
         \hspace{1.2cm}$8\times8$
      \end{flushleft}
       }
      \includegraphics[width=0.95\linewidth]{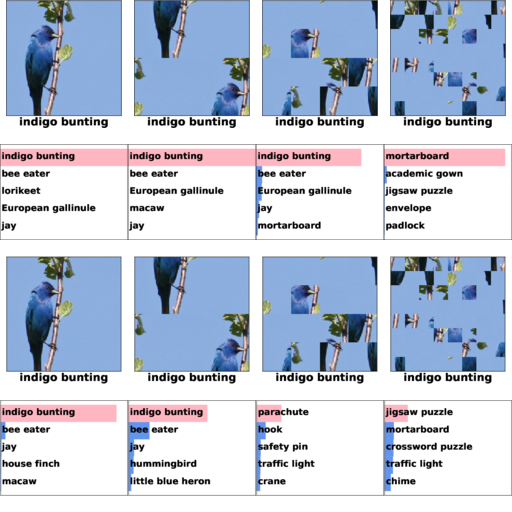}

   \end{subfigure}
   
   \caption{
       Qualitative examples showing the strong \emph{texture} bias of standard CNNs (here, ResNet) and the strong \emph{shape} bias of adversarially-robust models (here, ResNet-R) as described in Sec.~\ref{sec:distorted_shape}.
       When patches are scrambled, ResNet-R confidence drops substantially and its \colorbox{pink!100}{\strut top-1 predicted} labels often change away from the original, correctly-predicted label, here, \class{rule} (left) and \class{indigo~bunting} (right).}
   \label{fig:resnet_top-5-scrambled}
\end{figure*}

We create shape-less images by dividing each ImageNet-CL image into a grid of $p\times p$ even patches where $p \in \{2, 4, 8\}$ and re-combining them randomly into a new ``scrambled'' version (Fig.~\ref{fig:unseen}c).
On average, over three grid types, we observed a larger accuracy drop in R models compared to S models, ranging from 1.6$\times$ to $2.04\times$ lower accuracy (Table~\ref{tab:adv_acc}d).
That is, \textbf{R models become substantially less accurate than S models when shape cues are removed by patch-shuffling}---another evidence for their exclusive reliance on shapes (instead of textures).
See Fig.~\ref{fig:resnet_top-5-scrambled} for the predictions of ResNet and ResNet-R on scrambled patches.

\begin{figure}[t]
   \centering
   \includegraphics[width=1.0\linewidth]{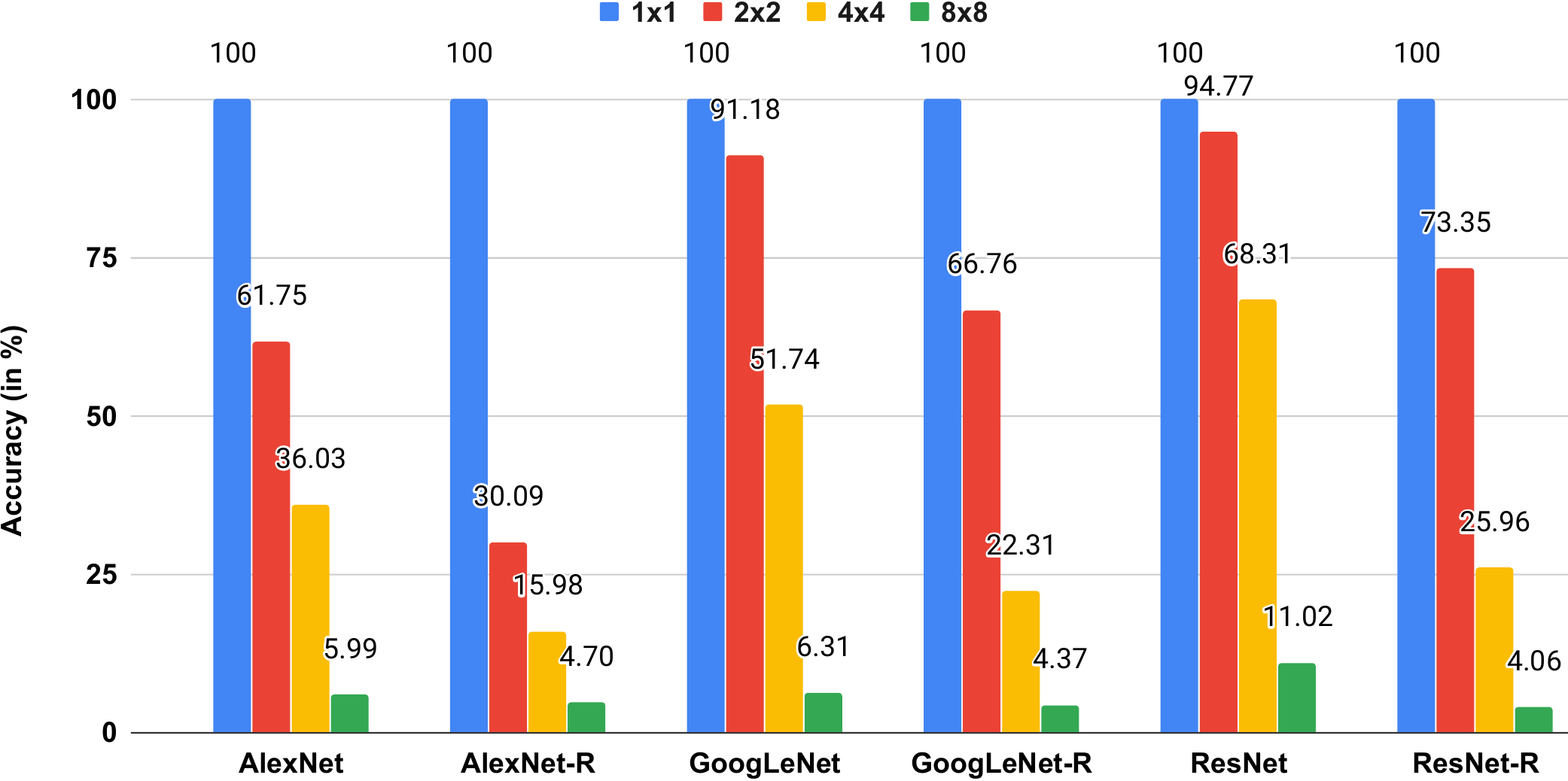}
   \caption{
       Standard CNNs substantially outperform R models on scrambled versions of the ImageNet-CL images due to their capability of recognizing images using textures.
      % See Fig.~\ref{fig:resnet_top-5-scrambled} for examples of scrambled images and their top-5 predictions from ResNet-R and ResNet (which achieves a remarkable accuracy of 94.77\%).\\
       Here, we report top-1 accuracy (\%) on the scrambled images whose original versions (ImageNet-CL) were correctly-labeled by both standard and R classifiers (hence, the 100\% for $1\times1$ blue bars).
   }
   \label{fig:accuracy_scrambled_images}
\end{figure}

\subsubsection{Performance on texture-less images}
\label{sec:texture-less}

% To understand their generalization capability that uses shape features, we followed

Following \cite{geirhos2018imagenet}, we test R models on three types of texture-less images where the texture is increasingly removed: (1) stylized ImageNet images where textures are randomly modified; (2) binary, black-and-white, \ie B\&W, images (Fig.~\ref{fig:unseen}e); and (3) silhouette images where the texture information is completely removed (Fig.~\ref{fig:unseen}f).

\subsec{Stylized ImageNet} 
To construct a set of stylized ImageNet images (see Fig.~\ref{fig:unseen}d), we take all ImageNet-CL images (Sec.~\ref{sec:networks_datasets}) and change their textures via a stylization procedure in \cite{geirhos2018imagenet}, which harnesses the style-transfer technique \citep{gatys2016image} to apply a random style to each ImageNet ``content'' image.
% That is, the stylized image sets are of the same sizes as the ImageNet-CL sets.

\subsec{B\&W images}
For all ImageNet-CL images, we use the same process described in \cite{geirhos2018imagenet} to generate silhouettes, but we do not manually select or modify the images. 
We use \citet{imagemagick} to binarize ImageNet images into B\&W images via the following steps:
\begin{verbatim}
convert image.jpeg image.bmp 
potrace -svg image.bmp -o image.svg
rsvg-convert image.svg > image.jpeg
\end{verbatim}

\subsec{Silhouette}
For all ImageNet-CL images, we obtain their segmentation maps via a PyTorch DeepLab-v2 model \citep{chen2017deeplab} pre-trained on MS COCO-Stuff.
We used the ImageNet-CL images that belong to a set of 16 COCO superclasses in \cite{geirhos2018imagenet} (\eg \class{bird}, \class{bicycle}, \class{airplane}).
When evaluating classifiers, an image is considered correctly labeled if its ImageNet predicted label is a subclass of the correct COCO superclass (Fig.~\ref{fig:unseen}f; \class{sandpiper} $\to$ \class{bird}).

\subsec{Results} Consistently, \textbf{on all three texture-less image sets, R models outperformed their S counterparts} (Table~\ref{tab:adv_acc}d--f)---a remarkable generalization capability, especially on B\&W and silhouette images where little to no texture information is available.

% \anh{Editing here}

% \clearpage
% \newpage

\subsection{What internal mechanisms make adversarially trained CNNs more robust than standard CNNs?}
\label{sec:R_representation}

\begin{figure*}[t]
   \centering
   \begin{subfigure}{0.49\linewidth}
      \centering
      \includegraphics[width=0.98\linewidth]{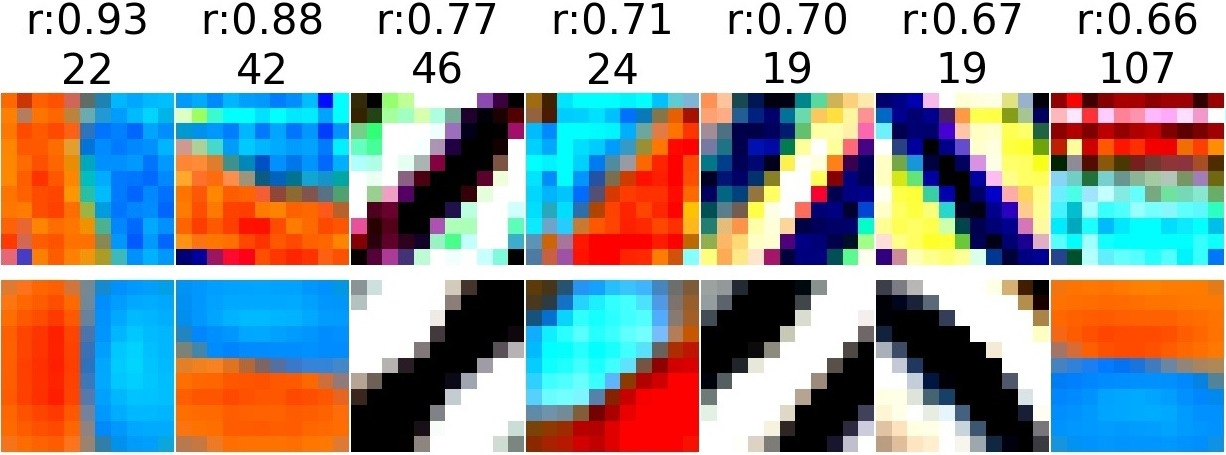}
      \caption{
         Standard filters (top) \& matching R filters (bottom)
    %       \vspace{0.3cm}
      }
      \label{fig:weight_smooth}
   \end{subfigure}
   % Original NeurIPS compact format
   \begin{subfigure}{0.49\linewidth}
      % \vspace*{-0.6cm}
      \centering
      {
          \small
         \begin{flushleft}
         \hspace{1.5cm}AlexNet
         \hspace{2.7cm}AlexNet-R
         \end{flushleft}
       }
      \includegraphics[width=0.95\linewidth]{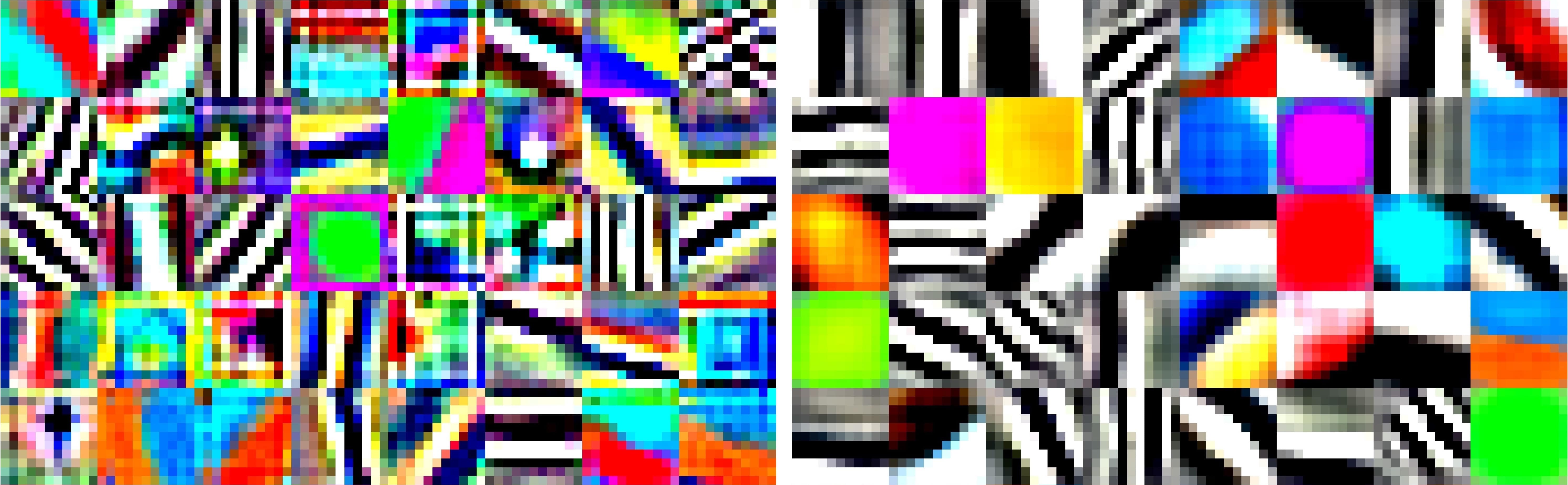}
      \caption{
      40 \layer{conv1} filters in AlexNet and AlexNet-R
      }
      \label{fig:cropped_alexnet_weights}
   \end{subfigure}

   \caption{
       \textbf{Left:} For each AlexNet \layer{conv1} filter (top row), we show the highest-correlated filter in AlexNet-R (bottom row), their Spearman rank correlation (e.g. \layer{r}:~\layer{0.93}) and the Total Variation (TV) difference (\eg \layer{22}) between the top kernel and the bottom. 
       Here, the TV differences are all positive i.e. AlexNet filters have higher TV.
       See Fig.\ref{fig:compare_corr_alexnet_weights} for full plot.
       \textbf{Right:} \layer{conv1} filters of AlexNet-R are smoother and less diverse than the counterparts.
       See Figs.~\ref{fig:full_conv1_weights} for GoogLeNet \& ResNet.
   }
   \label{fig:alexnet_smooth_weights}
\end{figure*}

\begin{table*}[ht]
   \caption{
   Mean total variation (TV) of the \layer{conv} layers of 6 models. 
   }
   \label{tab:total_variance}
   \def\arraystretch{1.1}%
   \begin{center}
      \begin{tabular}{l|c|c|c|c|c|c}
         \hline
         & AlexNet & AlexNet-R & GoogLeNet & GoogLeNet-R & ResNet & ResNet-R \\
         \hline
%        ImageNet(old) & \textbf{56.52} & 39.83 & \textbf{68.86} & 50.94 & \textbf{75.59} & 56.25 \\ \hline
         Mean TV & 110.20 & \textbf{63.59} & 36.53 & \textbf{22.79} & 18.35 & 19.96 \\ \hline
      \end{tabular}
   \end{center}
\end{table*}

We have shown that after adversarial training, R models are more robust than S models on new adversarial examples generated for these pre-trained models via PGD attacks (Table~\ref{tab:adv_acc}b).
Furthermore, on non-adversarial, high-frequency images, R models may also outperform S models (Table~\ref{tab:imagenet_c}a; AlexNet-R) \citep{yin2019fourier,gilmer2019adversarial}.

We aim to understand the internal mechanisms that make R CNNs more robust to high-frequency noise by analyzing the networks at the weight (Sec.~\ref{sec:smooth_tv}) and neuron (Sec.~\ref{sec:netdissect_robust_difference}) levels.

\subsubsection{Weight level: Smooth filters to block pixel-wise noise}
\label{sec:smooth_tv}

\subsec{Smoother filters}
To explain this phenomenon, we visualized the weights of all 64 \layer{conv1} filters (11$\times$11$\times$3), in both AlexNet and AlexNet-R, as RGB images.
We compare each AlexNet \layer{conv1} filter with its nearest \layer{conv1} filter (via Spearman rank correlation) in AlexNet-R. 
Remarkably, R filters appear qualitatively much smoother than their counterparts (Fig.~\ref{fig:weight_smooth}).
The R filter bank is also less diverse, \eg R edge detectors are often black-and-white in contrast to the colorful AlexNet edges (Fig.~\ref{fig:cropped_alexnet_weights}).
A similar contrast was also seen for the GoogL-eNet and ResNet models (Fig.~\ref{fig:full_conv1_weights}).

We also quantify the smoothness, in total variation (TV) \citep{rudin1992nonlinear}, of the filters of all 6 models (Table~\ref{tab:total_variance}) and found that, on average, the filters in R networks are much smoother.

For example, the mean TV of GoogLeNet-R is about 1.5 times smaller than that of GoogLeNet. 

In almost all layers, R filters are smoother than S filters (Fig.~\ref{fig:Appendix_kernel_tv}).

\subsec{Blocking pixel-wise noise}

We hypothesize that the smoothness of filters makes R classifiers more robust against noisy images. 

To test this hypothesis, we computed the total variation of the channels across 5 \layer{conv} layers when feeding ImageNet-CL images and their noisy versions (Fig.~\ref{fig:unseen}c; ImageNet-C Level 1 additive noise $\sim N(0, 
0.08)$) to S and R models.

\begin{figure*}[hbtp]
   \centering
   \begin{subfigure}{0.31\linewidth}
      \centering
      \includegraphics[width=1.0\linewidth]{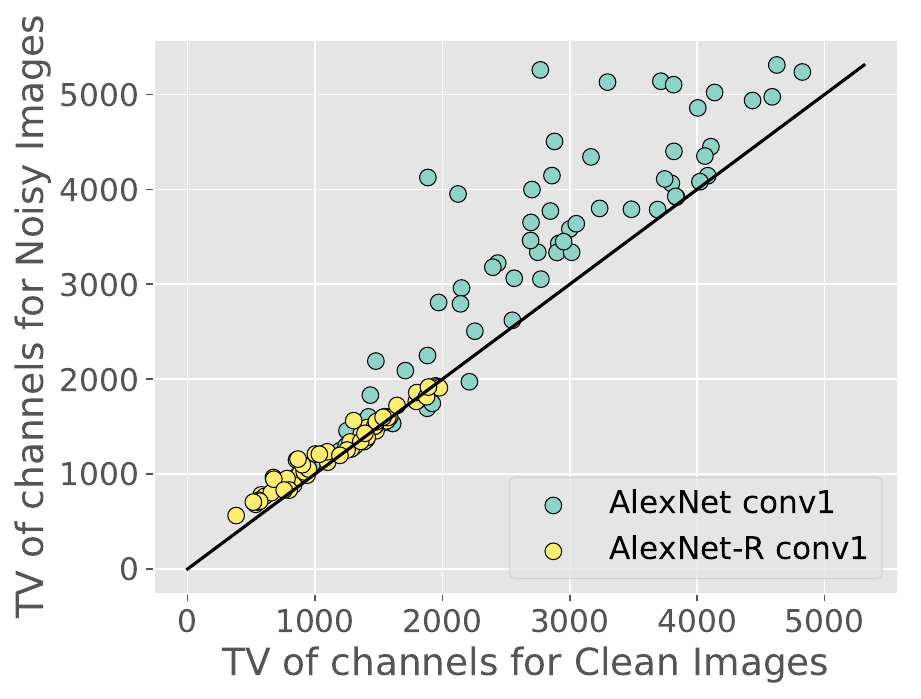}
      \caption{64 \layer{conv1} channels
      }
      \label{fig:tv_conv1}
   \end{subfigure}
   \begin{subfigure}{0.31\linewidth}
      \centering
      \includegraphics[width=1.0\linewidth]{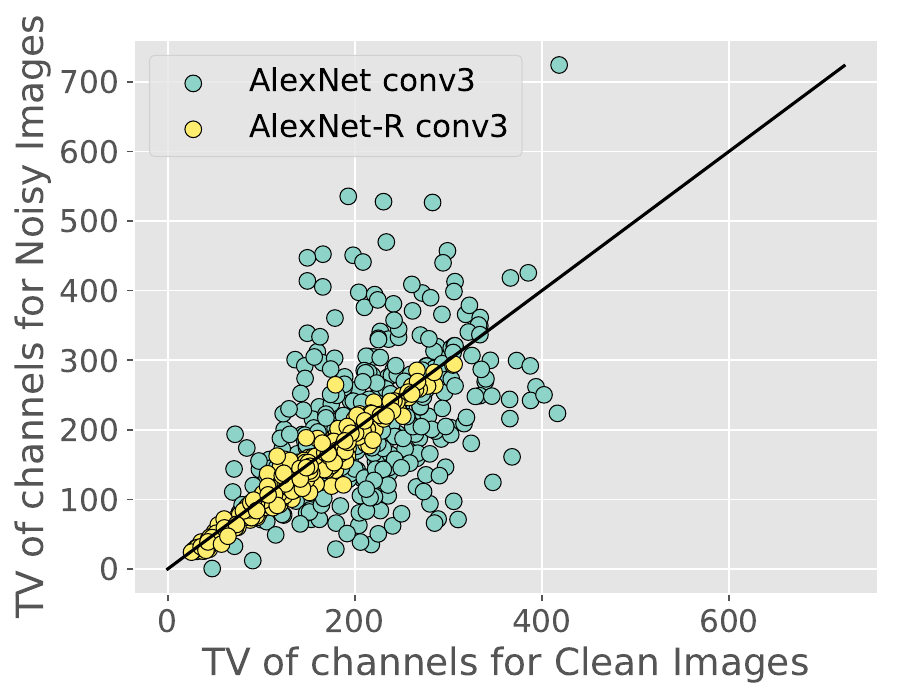}
      \caption{384 \layer{conv3} channels
      }
      \label{fig:tv_conv3}
   \end{subfigure}
   \begin{subfigure}{0.31\linewidth}
      \centering
      \includegraphics[width=1.0\linewidth]{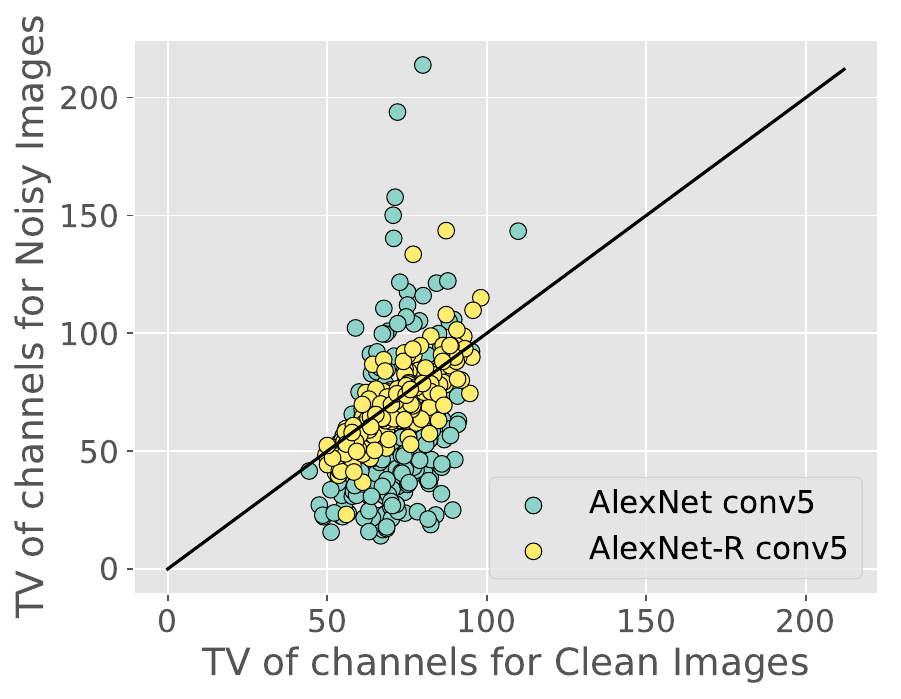}
      \caption{256 \layer{conv5} channels
      }
      \label{fig:tv_conv5}
   \end{subfigure}
   \hspace{0.03\linewidth}

   \caption{
       In each subpanel, one point shows the mean Total Variation (TV) of one channel when running clean ImageNet-CL images and their noisy versions through AlexNet (\mycircle{BlueGreen}) or AlexNet-R (\mycircle{yellow}).
       R channels have similar TV before and after adding noise, suggesting that \layer{conv1} kernels filter out the added noise.
       In higher layers (\layer{conv3} and \layer{conv5}), R channels are consistently more invariant to the input noise than S channels (\mycircle{yellow} dots are clustered around the diagonal line while \mycircle{BlueGreen} dots have higher variance).
       See Fig.~\ref{fig:si_clean_noisy_tv}
       for the same scatter plot (a) for all five layers.
       }
   \label{fig:alexnet_tv}
\end{figure*}

At \layer{conv1}, the smoothness of R activation maps remains almost unchanged before and after noise addition (Fig.~\ref{fig:tv_conv1};

yellow circles are on the diagonal line).
In contrast, the \layer{conv1} filters in standard AlexNet allow Gaussian noise to pass through, yielding larger-TV channels (Fig.~\ref{fig:tv_conv1}; blue circles are mostly above the diagonal).
That is, the smooth filters in \textbf{R models indeed can filter out pixel-wise Gaussian noise despite that R models were not explicitly trained on this image type}!
Furthermore, AlexNet-R and ResNet-R consistently perform better in different noise distortions (Table ~\ref{tab:imagenet_c} (a)) compared to their S models.

In higher layers, it is intuitive that the pixel-wise noise added to the input image might not necessarily cause activation maps, in both S and R networks, to be noisy because higher-layered units detect more abstract concepts.
However, interestingly, we still found that R channels to have consistently less mean TV (Fig.~\ref{fig:tv_conv3}--c).

Our result suggests that most of the de-noising effects take place at lower layers (which contain more generic features) instead of higher layers (which contain more task-specific features).

\subsubsection{Neuron level: Robust neurons prefer lower-level and fewer inputs}
\label{sec:netdissect_robust_difference}

Here, via NetDissect framework, we wish to characterize how adversarial training changed the hidden neurons in R networks to make R classifiers more adversarially robust.

\subsec{Network Dissection} (hereafter, NetDissect) is a common framework for quantifying the functions of a neuron by computing the Intersection over Union (IoU) between each activation map (\ie channels) and the human-annotated segmentation maps for the same input images.

That is, each channel is given an IoU score per human-defined concept (\eg \class{dog} or \class{zigzagged}) indicating its accuracy in detecting images of that concept.
A channel is tested for its accuracy on all $\sim$1,400 concepts,

which span across six coarse categories: \class{object}, \class{part}, \class{scene}, \class{texture}, \class{color}, and \class{material} \citep{bau2017network} (c.f. Fig.~\ref{fig:alexnet_top5_concept} for example NetDissect images in \class{texture} and \class{color} concepts).

Following \cite{bau2017network}, we assign each channel $C$ a main functional label \ie the concept that $C$ has the highest IoU with.
In both S and R models, we ran NetDissect on all 1152, 5808, and 3904 channels from, respectively, 5, 12, and 5 main convolutional layers (post-ReLU) of the AlexNet, GoogLeNet, and ResNet-50 architectures (c.f. Sec.~\ref{sec:list_netdissect_layers} for more details of layers used).

\begin{figure}[t]
   \centering
   \includegraphics[width=1.0\linewidth]{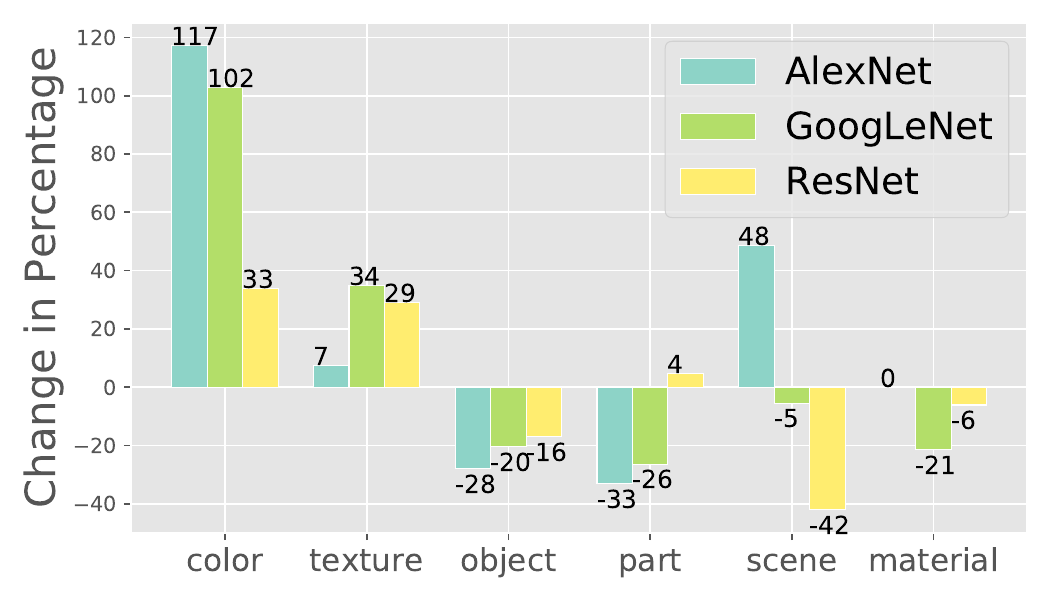}
   \caption{
       Total channel increases (\%) in R models. For all three architectures, the numbers of NetDissect \class{color} and \class{texture} detectors in R models increase, \eg by 117\% and 7\%, respectively, for AlexNet, while the number of \class{object} units decreases by 28\%.
   See Fig.~\ref{fig:alexnet_object_color} for layer-wise plots for AlexNet in other object and color category.
   }
   \label{fig:global_difference_plot}
\end{figure}

\subsec{Shift to detecting more low-level features \ie colors and textures}
We found a consistent trend---adversarial training resulted in substantially more filters that detect colors and textures (\ie in R models) in exchange for fewer object and part detectors.

\begin{figure*}[hbtp]
   \centering
   \begin{subfigure}{0.40\linewidth}
   \centering
   \includegraphics[width=1.0\linewidth]{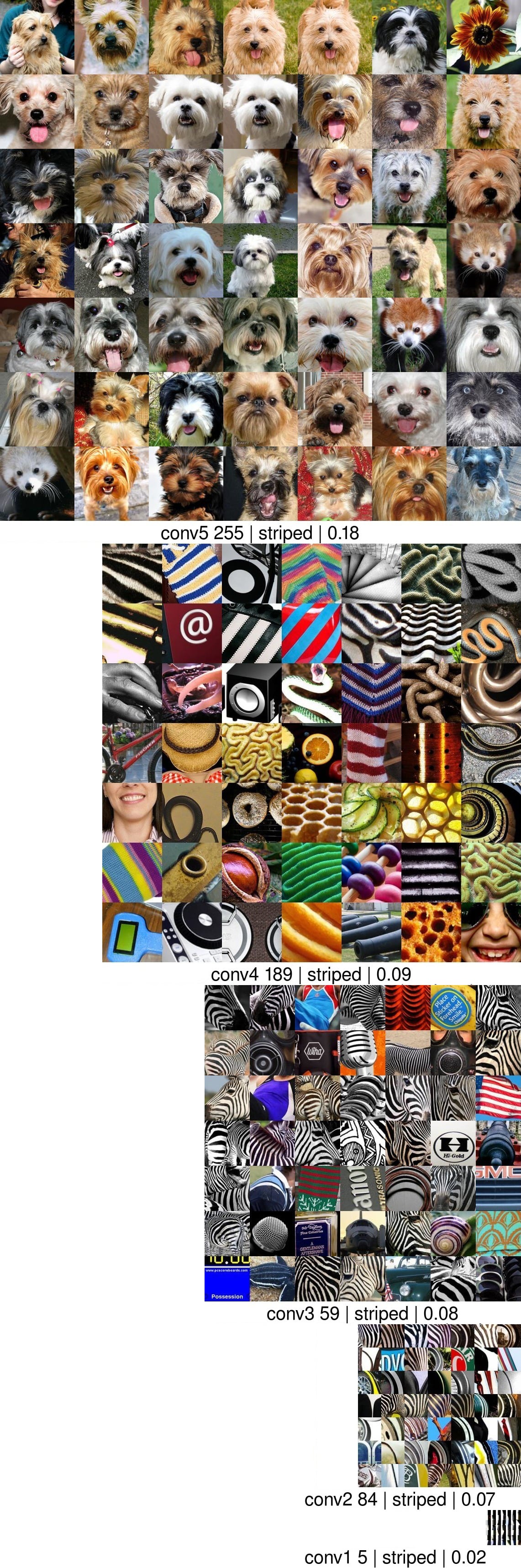}
   \caption{
       AlexNet
   }
   \label{fig:alexnet_top49_striped}
   \end{subfigure}
   \begin{subfigure}{0.40\linewidth}
      \centering
   \includegraphics[width=1.0\linewidth]{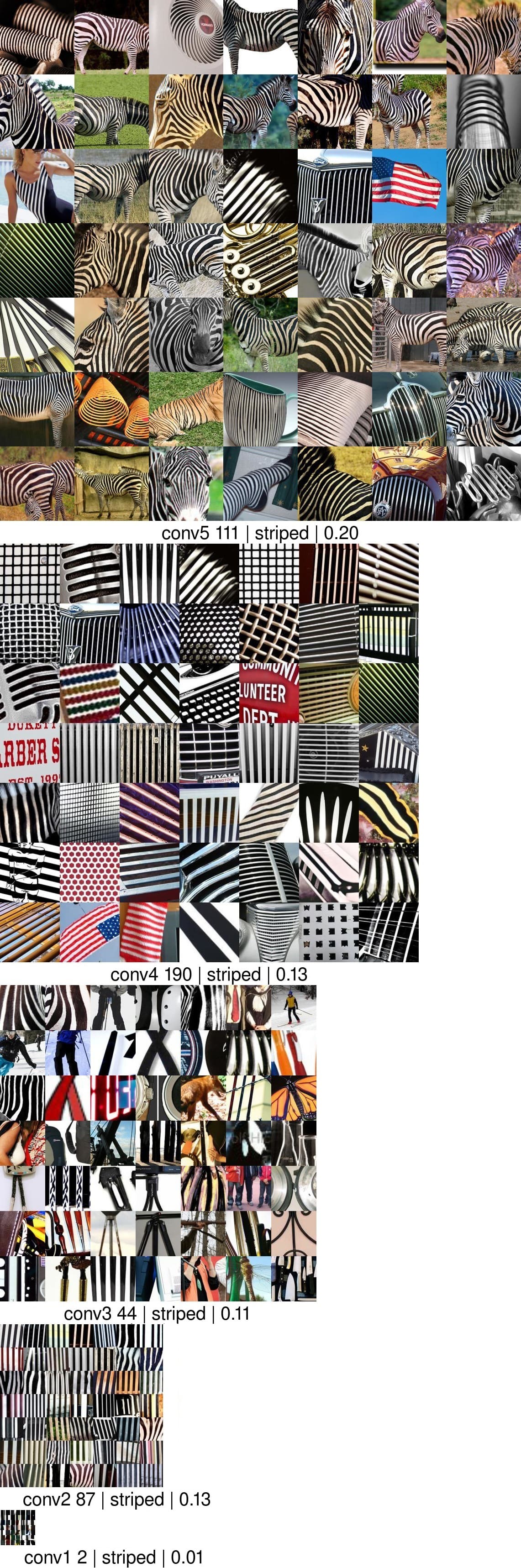}
   \caption{
       AlexNet-R
   }
   \label{fig:alexnet_r_top49_striped}
   \end{subfigure}
   \caption{
       Each 7$\times$7 grid shows the top-49 training-set images that highest activate the center unit in a channel.
       Each column shows five highest-IoU \class{striped} concept channels, each from one AlexNet's \layer{conv} layer in their original resolutions.
       From top to bottom, AlexNet-R (b) consistently preferred striped patterns, \ie, edges (\layer{conv1}), vertical bars (\layer{conv2}), tools, to grids and zebra (\layer{conv5}).
       In contrast, AlexNet \class{striped} images (a) are much more diverse, including curly patterns (\layer{conv4}) and dog faces (\layer{conv5}). 
   }
   \label{fig:top49_striped}
\end{figure*}

For example, throughout the same GoogLeNet architecture, we observed a 102\% and a 34\% increase of color and texture detectors, respectively, in the R model, but a 20\% and a 26\% fewer object and part detectors, compared to the S model (c.f. Fig.~\ref{fig:global_difference_plot}).
After adversarial training, $\sim$11\%, 15\%, and 10\% of all hidden neurons (in the tested layers) in AlexNet, GoogLeNet, and ResNet, respectively, shift their roles to detecting lower-level features (\ie textures and colors) instead of higher-level features (see feature visualizations in Fig.~\ref{fig:top49_striped}).

Across three architectures, the increases in \class{texture} and \class{color} channels are often larger in higher layers.
The largest functional shifts appearing in higher layers can be because the higher-layered units are more task-specific ~\citep{nguyen2016synthesizing,yosinski2014transferable}.

\begin{figure}[hbtp]
   \centering
   \includegraphics[width=1\linewidth]{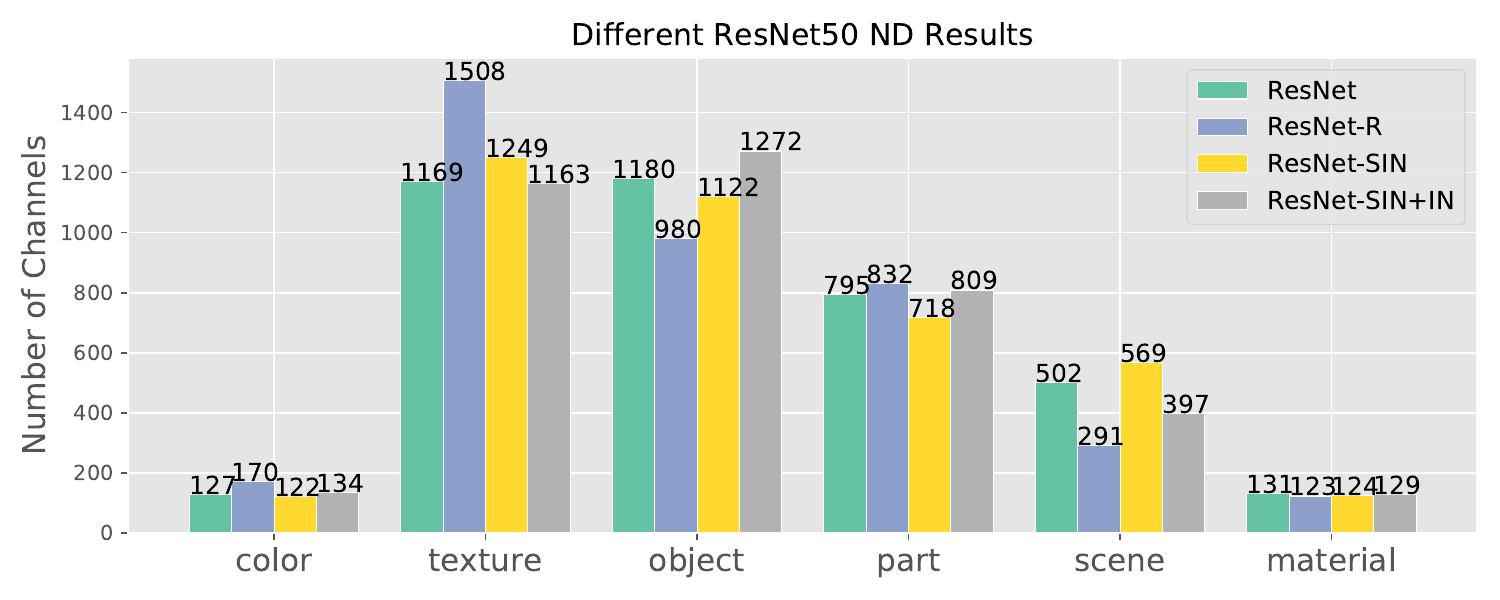}
   \caption{
%  \todo{smaller y-axis labels, smaller legend}
   Each column shows the number of channels (in a ResNet-50 model) categorized into one of NetDissect categories. 
   For example, the standard ImageNet-trained \colorbox{BlueGreen!50}{\strut ResNet}
   has 127 \class{color} detectors (leftmost column).
    Here, we compare the neural functions among four different ResNet-50 models trained differently: (1) \colorbox{BlueGreen!50}{\strut ResNet} was trained on ImageNet; (2) \colorbox{blue!20}{\strut ResNet-R} was trained via PGD-adversarial training; (3) \colorbox{yellow!100}{\strut ResNet-SIN} was trained on texture-removed ImageNet from \citep{geirhos2018imagenet};
    and (4) \colorbox{gray!30}{\strut ResNet-SIN+IN}
    was a ResNet-SIN that was then finetuned on ImageNet.
    Training on texture-removed ImageNet or adversarial examples consistently produces CNNs that are (a) heavily shape-biased; (b) contain more low-level, \class{texture} features; (c) fewer high-level, \class{object} detectors compared to training on ImageNet which produces texture-biased CNNs (ResNet and ResNet-SIN+IN).
      } 
   \label{fig:resnet_variance}
\end{figure}

\subsec{Consistent findings with ResNet CNNs trained on Stylized ImageNet}
We also compare the shape-biased ResNet-R with ResNet-SIN, \ie a ResNet-50 trained exclusively on stylized ImageNet images where textures are removed via stylization  \citep{geirhos2018imagenet}.
ResNet-SIN also has a strong shape bias of 81.37\%.\footnote{\class{model\_A}~in~\url{https://github.com/rgeirhos/texture-vs-shape/}. See Table 4 in \cite{geirhos2018imagenet}.}
Interestingly, similar to ResNet-R, ResNet-SIN also has more low-level feature detectors (colors and textures) and fewer high-level feature detectors (objects and parts) than the vanilla ResNet (Fig.~\ref{fig:resnet_variance}).
In contrast, finetuning this ResNet-SIN on ImageNet remarkably changes the model to be \emph{texture-biased} (at a 79.7\% texture bias) and to contain fewer \class{texture} and more \class{object} and \class{part} units (Fig.~\ref{fig:resnet_variance}; ResNet-SIN+IN vs. ResNet-SIN).

That is, training or finetuning on ImageNet tend to cause CNNs to be more texture-biased and contain more high-level features (\ie detecting objects and parts).
In contrast, training on adversarial examples or texture-distorted images cause CNNs to focus more on shapes and learn more generic, low-level features.
% \anh{Editing here}

\begin{figure*}[t]
   \centering
   \includegraphics[width=0.7\linewidth]{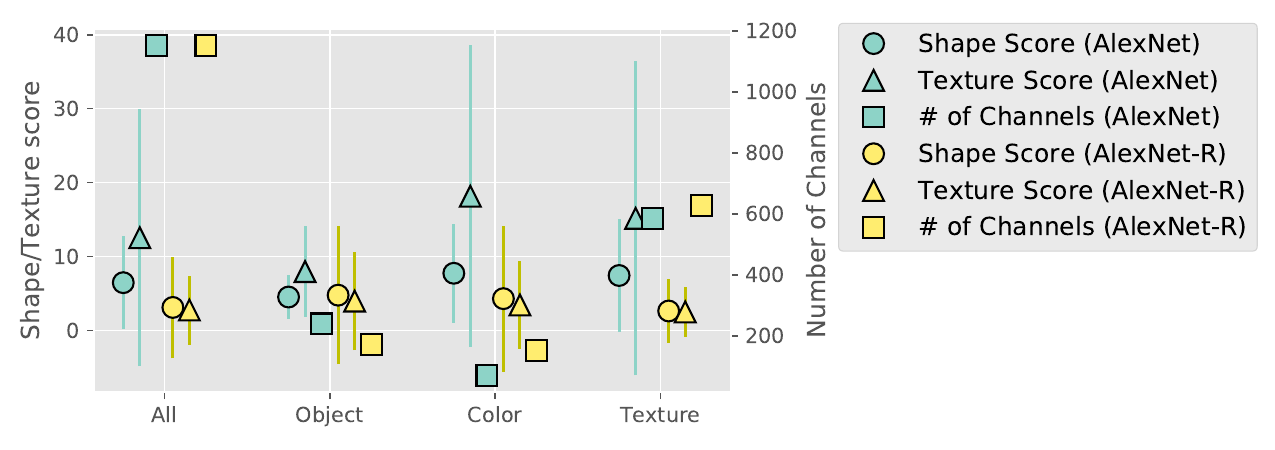}
   \caption{
       Shape \& Texture score of AlexNet \& AlexNet-R. The average Shape (\mycircle{BlueGreen}) and Texture (\mytriangle{BlueGreen}) scores over all channels in the entire network (``All'') or in a NetDissect category (``Object'', ``Color'', and ``Texture'').
   While AlexNet-R has more \class{color} and \class{texture} channels (\mysquare{yellow} above \mysquare{BlueGreen}), these R channels are not heavily shape- or texture-biased.
   In contrast, the corresponding channels in AlexNet are heavily texture-biased (\mytriangle{BlueGreen} is almost 2$\times$ of \mycircle{BlueGreen}).
   }
   \label{fig:shap_texture_score_alexnet}
\end{figure*}

\begin{figure*}[htbp]
   \centering
   \includegraphics[width=1\linewidth]{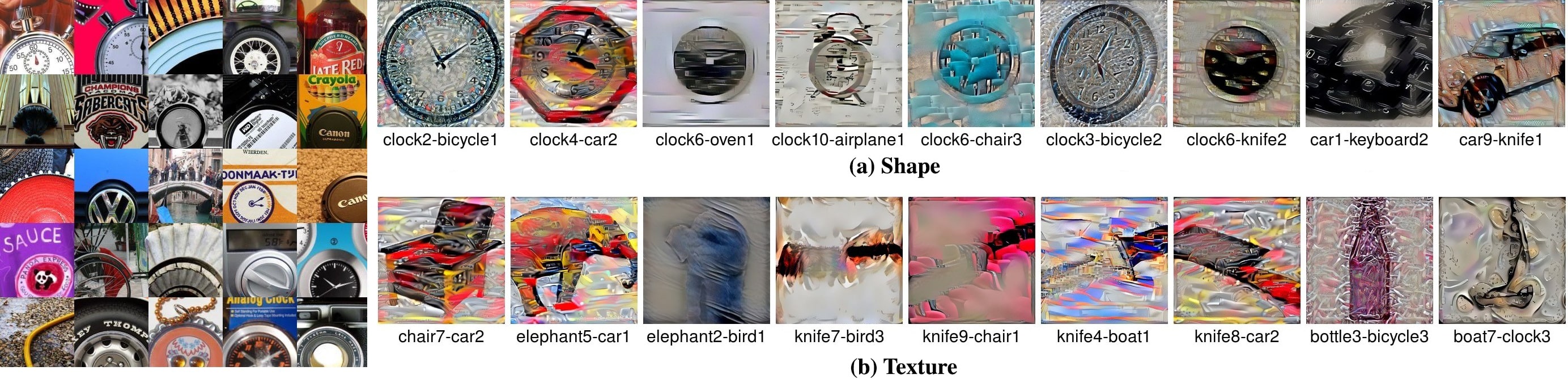}
   \caption{
      \textbf{Left:} Top-25 highest-activation images of the AlexNet unit~\layerunit{conv4}{19}, which has a NetDissect label of \class{spiralled} under \class{texture} category.
      The unit prefers circular patterns including car wheels and clock.
      \textbf{Right:} 
      Example \emph{cue-conflict} images originally labeled by AlexNet as \class{shape} (top) or \class{texture} (bottom) but that were given a different label after the unit~\layerunit{conv4}{19} is ablated.
      For example, ``clock2-bicycle1'' is a cue-conflict image that has the shape of a clock and the texture of a bicycle.
      Qualitatively, the unit helps AlexNet detect clocks and cars using shapes (top) and reddish pink cars, birds, chairs, and bicycles using textures  (bottom).
      The unit has Shape and  Texture scores of 18 and 22, respectively.
      See Fig.~\ref{fig:Appendix_top49_cue_conflict_visulization_1} \& Fig.~\ref{fig:Appendix_top49_cue_conflict_visulization_7} for more examples.
      } 
   \label{fig:st_visualization}
\end{figure*}

\subsec{Shift to detecting simpler objects}
Analyzing the concepts in the \class{object} category where we observed largest changes in channel count, we found evidence that neurons change from detecting complex to simpler objects.
That is, for each NetDissect concept, we computed the difference in the numbers of channels between the S and R model.
% The concepts with the largest differences suggest that the neurons change from detecting complex objects to simpler objects as the result of adversarial training.
In the same \class{object} category, AlexNet-R model has substantially fewer channels detecting complex concepts \eg $-30$ \layer{dog}, $-13$ \layer{cat}, and $-11$ \layer{person} detectors (Fig.~\ref{fig:alexnet_object_diff}; rightmost columns), compared to the standard network.
In contrast, the R model has more channels detecting simpler concepts, \eg $+40$ \layer{sky} and $+12$ \layer{ceiling} channels (Fig.~\ref{fig:alexnet_object_diff}; leftmost columns).
The top-49 images that highest-activated R units across five \layer{conv} layers also show their strong preference for simpler backgrounds and objects
(Fig.~\ref{fig:top49_striped}).

% (Figs.~\ref{fig:appendix_conv1}--\ref{fig:appendix_conv5}).

% See Figs.~\ref{fig:appendix_conv1}-\ref{fig:appendix_conv5} for comparisons of top-49 images of units in all five \layer{conv} layers.

% R model has substantially less complex channels \eg $-30$ \layer{dog}, $-13$ \layer{cat}, and $-11$ \layer{person} detectors (Fig.~\ref{fig:alexnet_object_diff}; rightmost columns).

% prefer more simpler objects than complex objects.
% For example, in the same \class{object} category, AlexNet-R model has $+40$ \layer{sky} and $+12$ \layer{ceiling} channels more than the standard model (Fig.~\ref{fig:alexnet_object_diff}; leftmost columns).
% In contrast, the 

% \vspace{-0.2cm}
% \begin{figure}[ht]
%  \centering
%  \includegraphics[width=1.0\linewidth]{images/Diff_object.pdf}
%  \caption{\todo{Write caption} object difference plot. Network-wise}
%  \label{fig:alexnet_object_diff_small}
% \end{figure}

% \vspace{-0.5cm}

% \subsubsection{Neuron level: Robust neurons detect fewer unique concepts}
% \label{sec:simpler_units}

\subsec{Shift to detecting fewer unique concepts}
The previous sections have revealed that neurons in R models often prefer images that are pixel-wise smoother (Sec.~\ref{sec:smooth_tv}) and of lower-level features (Sec.~\ref{sec:netdissect_robust_difference}), compared to S neurons.
Another important property of the complexity of the function computed at each neuron is the diversity of types of inputs detected by the neuron \citep{nguyen2016multifaceted,nguyen2019understanding}.
Here, we compare the diversity score of NetDissect concepts detected by units in S and R networks.
For each channel $C$, we calculated a diversity score, \ie the number of unique concepts that $C$ detects with an IoU score $\geq 0.01$.
% A channel that detects a more diverse set of concepts is considered more complex.
% For both networks, channels in higher layers are often more complex than those in lower layers (Fig.~\ref{fig:alexnet_diversity_plot}; increasing trend).
% See the increasing diversity in the highest-activation images in 
% Fig.~\ref{fig:striped_top49_progression} for example diversity of highest-activation images of channels across five layers.

% \chirag{In Fig.~\ref{fig:concept_top49} we observe that the \class{sky} and \class{striped} concepts in AlexNet are also activated for animal images.}

Interestingly, on average, an R unit fires for 1.16 times fewer unique concepts than an S unit (22.43 vs. 26.07; c.f. Fig.~\ref{fig:alexnet_diversity_plot}).
% are consistently less multi-functional than S channels in all 5 \layer{conv} layers of AlexNet architecture.
% For example, at \layer{conv5}, one AlexNet unit detects, on average, 38 concepts while that mean diversity score for AlexNet-R is only 32 (Fig.~\ref{fig:alexnet_diversity_plot}; \layer{conv5}).
Similar trends were observed in ResNet (Fig.~\ref{fig:resnet_diversity_plot}).
Qualitatively comparing the highest-activation training-set images by the highest-IoU channels in both networks, for the same most-frequent concepts (\eg \layer{striped}), often confirms a striking difference: R units prefer a less diverse set of inputs (Fig.~\ref{fig:top49_striped}).
As R hidden units fire for fewer concepts, \ie significantly fewer inputs, the space for adversarial inputs to cause R models to misbehave is strictly smaller.

\subsection{Which neurons are important for shape-based or texture-based image classification?}
\label{sec:R_net_shift_to_shape}
To understand how the found changes in R neurons (Sec.~\ref{sec:R_representation}) relate to the shape bias of R CNNs (Sec.~\ref{sec:shapes_or_textures}), here, we zero out every channel, one at a time, in S and R networks and measure the performance drop in recognizing shapes and textures from cue-conflict images.

\subsec{Shape \& Texture scores}
For each channel, we computed a ``Shape score'', \ie the number of images originally correctly labeled into the \class{shape} class by the network but that, after the ablation, are labeled differently (examples in Fig~\ref{fig:st_visualization}a--b).
Similarly, we computed a Texture score per channel.
The Shape and Texture scores quantify the importance of a channel in image classification using shapes and textures, respectively.

First, we found that the \textbf{channels labeled \class{texture} by NetDissect are not only important to texture-based but also shape-based classification}.
That is, on average, zero-ing out these channels caused non-zero Texture and Shape scores

(Fig.~\ref{fig:shap_texture_score_alexnet}; Texture 
\mycircle{yellow} and \mytriangle{yellow} are above 0).
See Fig.~\ref{fig:st_visualization} for an example of \class{texture} channels with high Shape and Texture scores.

This result is aligned with the fact that R networks consistently have more \class{texture} units (Fig.~\ref{fig:global_difference_plot}) but are shape-biased (Sec.~\ref{sec:shapes_or_textures}).

Second, the \class{texture} units are, as expected, highly texture-biased in AlexNet (Fig.~\ref{fig:shap_texture_score_alexnet} Texture; \mytriangle{BlueGreen} is almost 2$\times$ of \mycircle{BlueGreen}).
However, surprisingly, those \class{texture} \textbf{units in AlexNet-R are neither strongly shape-biased nor texture-biased} (Fig.~\ref{fig:shap_texture_score_alexnet}; Texture \mycircle{yellow} $\approx$ \mytriangle{yellow}).
That is, across all three groups of the \class{object}, \class{color}, and \class{texture}, \textbf{R neurons appear mostly to be generalist, low-level feature detectors}.
This generalist property might be a reason for why R networks are more effective in transfer learning than S networks \citep{salman2020adversarially}.

Finally, the contrast above between the texture bias of S and R channels (Fig.~\ref{fig:shap_texture_score_alexnet}) reminds researchers that the single semantic label assigned by NetDissect to each neuron is not describing a full picture of what the neuron does and how it helps in downstream tasks.
To the best of our knowledge, this is the first work to align the NetDissect and cue-conflict frameworks to study how individual neurons contribute to the generalizability and shape bias of the entire network.

\section{Related Work}
% \peijie{Maybe we could move the related work after introduction to resolve reviewer1's reference concern of missing citations of \cite{zhang2019interpreting}, we already cite it here.}

\subsec{Shape bias} 
On smaller-scaled datasets \eg CIFAR-10, \cite{zhang2019interpreting} also found that R networks rely heavily on shapes (instead of textures) to classify images.
However, such shape bias may not necessarily generalize to high-dimensional, large-scale datasets such as ImageNet.
Different from \cite{zhang2019interpreting}, here, we study networks trained on ImageNet and its variants (\eg Stylized ImageNet).
To the best of our knowledge, we are the first to reveal (1) the shape bias of ImageNet-trained R networks; (2) the roles of neurons in R networks; and (3) their internal robustification mechanisms.
We also found that our adversarially-trained CNNs and the CNNs trained on texture-removed images by \cite{geirhos2018imagenet} both similar have a strong shape bias and contain simpler and more generic features than standard ImageNet-trained CNNs.

\subsec{Simplicity bias} Deep neural networks tend to prioritize learning simple patterns that are common across the training set \citep{arpit2017closer}. 
Furthermore, deep ReLU networks often prefer learning simple functions \citep{valle-perez2018deep,de2019random}, specifically low-frequency functions \citep{rahaman2019on}, which are more robust to random parameter perturbations.
Along this direction, here, we have shown that R networks (1) have smoother weights (Sec.~\ref{sec:smooth_tv}), (2) prefer even simpler and fewer inputs (Sec.~\ref{sec:netdissect_robust_difference})
than standard deep networks---\ie R networks represent even simpler functions.
% and (4) detecting both shape and texture at the same time(Sec.~\ref{sec:R_net_shift_to_shape}).
Such simplicity biases are consistent with the fact that gradient images of R networks are much smoother \citep{tsipras2018robustness} and that R classifiers act as a strong image prior for image synthesis \citep{santurkar2019image}.

\subsec{Robustness} Each R neuron computing a more restricted function than an S neuron (Sec.~\ref{sec:netdissect_robust_difference}) implies that R models would require more neurons to mimic a complex S network.
This is consistent with recent findings that adversarial training requires a larger model capacity \citep{xie2020intriguing}.
In Sec.~\ref{sec:smooth_tv}, we found that adversarial training produces network that are robust to images with additive Gaussian noise.
Interestingly, \cite{ford2019adversarial} found the reverse is also true: Training CNNs with additive Gaussian noise can improve the performance on adversarial examples.

While AdvProp did not yet show benefits on ResNet \citep{xie2020adversarial}, it might be interesting future work to find out whether EfficientNets trained via AdvProp also have shape and simplicity biases.
Furthermore, simplicity biases may be incorporated as regularizers into future training algorithms to improve model robustness.
For example, encouraging filters to be smoother might improve robustness to high-frequency noise.

Also aligned with our findings, \cite{rozsa2019improved} found that explicitly narrowing down the non-zero input regions of ReLUs can improve adversarial robustness.

\section{Discussion and Conclusion}

We found that R networks heavily rely on shape cues in contrast to S networks.
One may fuse an S network and a R network (two channels, one uses texture and one uses shape) into a single, more robust, interpretable ML model.
That is, such model may (1) have better generalization on OOD data than S or R network alone and (2) enable an explanation to users on what features a network uses to label a given image.

Our study on how individual hidden neurons contribute to the R network shape preference (Sec.~\ref{sec:R_net_shift_to_shape})  revealed that texture-detector units are equally important to the texture-based and shape-based recognition. This is in contrast to a common hypothesis that texture detectors should be exclusively only useful to texture-biased recognition. Our surprising finding suggests that the categories of stimuli in the well-known Network Dissection \citep{bau2017network} need to be re-labeled and also extended with low-frequency patterns \eg single lines or silhouettes in order to more accurately quantify hidden representations.

It might be interesting future work to improve model performance by (1) training them jointly on adversarial examples and texture-less images; and (2) adding smoothness prior to network gradients and filters.
Also, adding a smoothness prior to the gradient images or filters of convolutional networks may improve their generalization capability.

In sum, our work has revealed several intriguing properties of adversarially-trained networks, providing insights for future designs of classifiers robust to out-of-distribution examples.

\section*{Acknowledgement}
AN is supported by the National Science Foundation under Grant No. 1850117 and donations from NaphCare Foundation, Adobe Research, and Nvidia.

\newpage
\section{Appendix}\label{appendix}
\appendix

\newcommand{\beginsupplementary}{
    \setcounter{table}{0}
    \renewcommand{\thetable}{A\arabic{table}}
    \setcounter{figure}{0}
    \renewcommand{\thefigure}{A\arabic{figure}}
    \setcounter{section}{0}
}
\beginsupplementary

\section{Convolutional layers used in Network Dissection analysis}
\label{sec:list_netdissect_layers}

For both standard and robust models, we ran NetDissect on 5 convolutional layers in AlexNet \citep{krizhevsky2012imagenet}, 12 in GoogLeNet \citep{szegedy2015going}, and 5 in ResNet-50 architectures \citep{he2016deep}.
For each layer, we use after-ReLU activations (if ReLU exists).

\paragraph{AlexNet} layers: \layer{conv1}, \layer{conv2}, \layer{conv3}, \layer{conv4}, \layer{conv5}.
Refer to these names in \citet{krizhevsky2012imagenet}.

\paragraph{GoogLeNet} layers: \layer{conv1}, \layer{conv2}, \layer{conv3}, \layer{inception3a},
\layer{inception3b}, \layer{inception4a}, \layer{inception4b}, \layer{inception4c}, \layer{inception4d}, \layer{inception4e}, 
\linebreak[4]
\layer{inception5a}, \layer{inception5b}

Refer to these names in PyTorch code \url{https://github.com/pytorch/vision/blob/master/torchvision/models/googlenet.py#L83-L101}.

\paragraph{ResNet-50} layers: \layer{conv1}, \layer{layer1}, \layer{layer2}, \layer{layer3}, \layer{layer4}

Refer to these names in PyTorch code \url{https://github.com/pytorch/vision/blob/master/torchvision/models/resnet.py#L145-L155}).

\section{Kernel smoothness visualization}
\label{sec:full_conv1_kernel_comp}

\begin{figure}[h]
   \centering
   \begin{subfigure}{1\linewidth}
      \centering
      \includegraphics[width=1\linewidth]{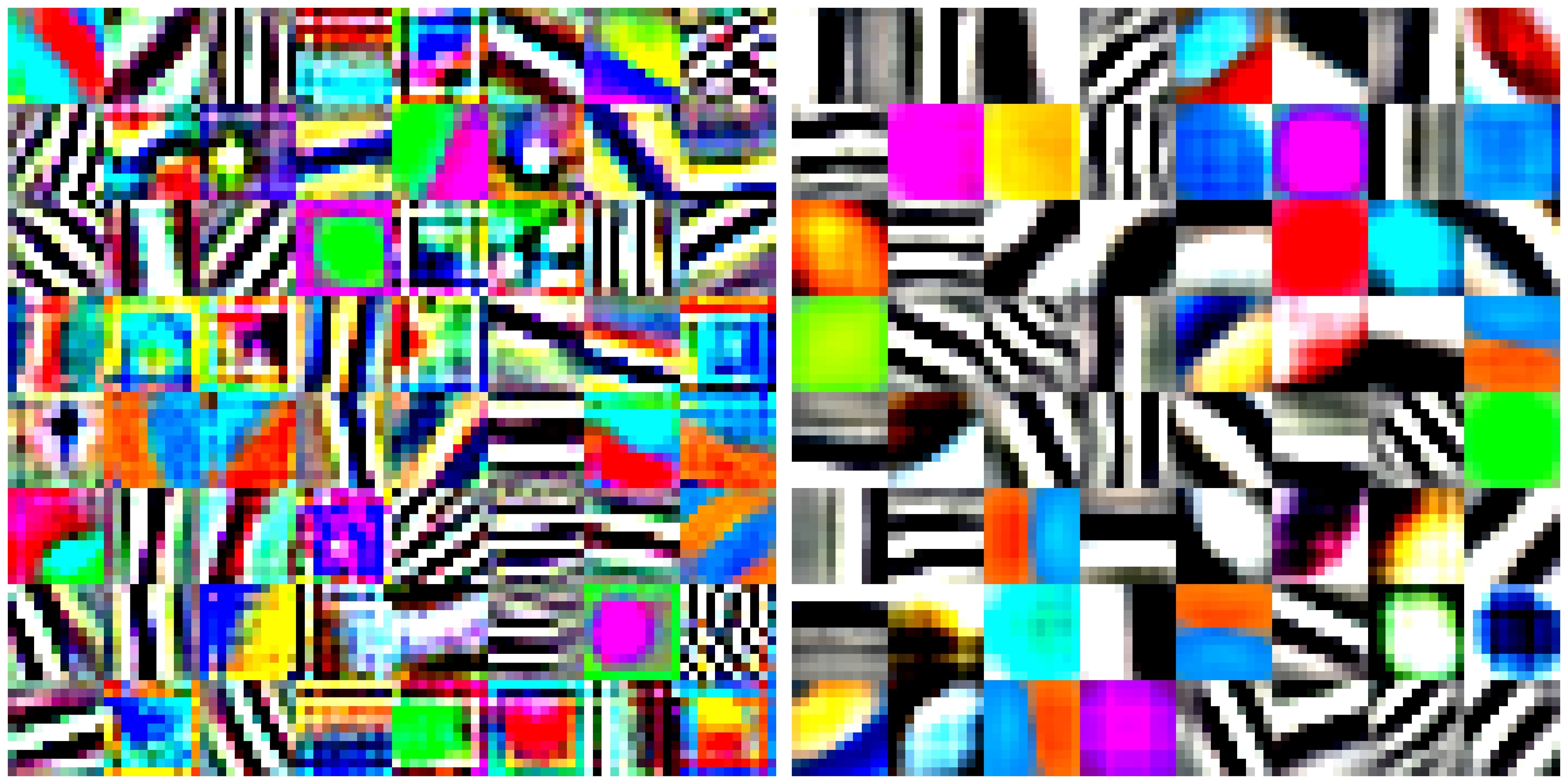}
      {
          \footnotesize
         \begin{flushleft}
            \hspace{1cm}AlexNet
            \hspace{1cm} 11$\times$11$\times$3
            \hspace{1cm}AlexNet-R
         \end{flushleft}
      }

   \end{subfigure}
   \begin{subfigure}{1\linewidth}
      \centering
      \includegraphics[width=1\linewidth]{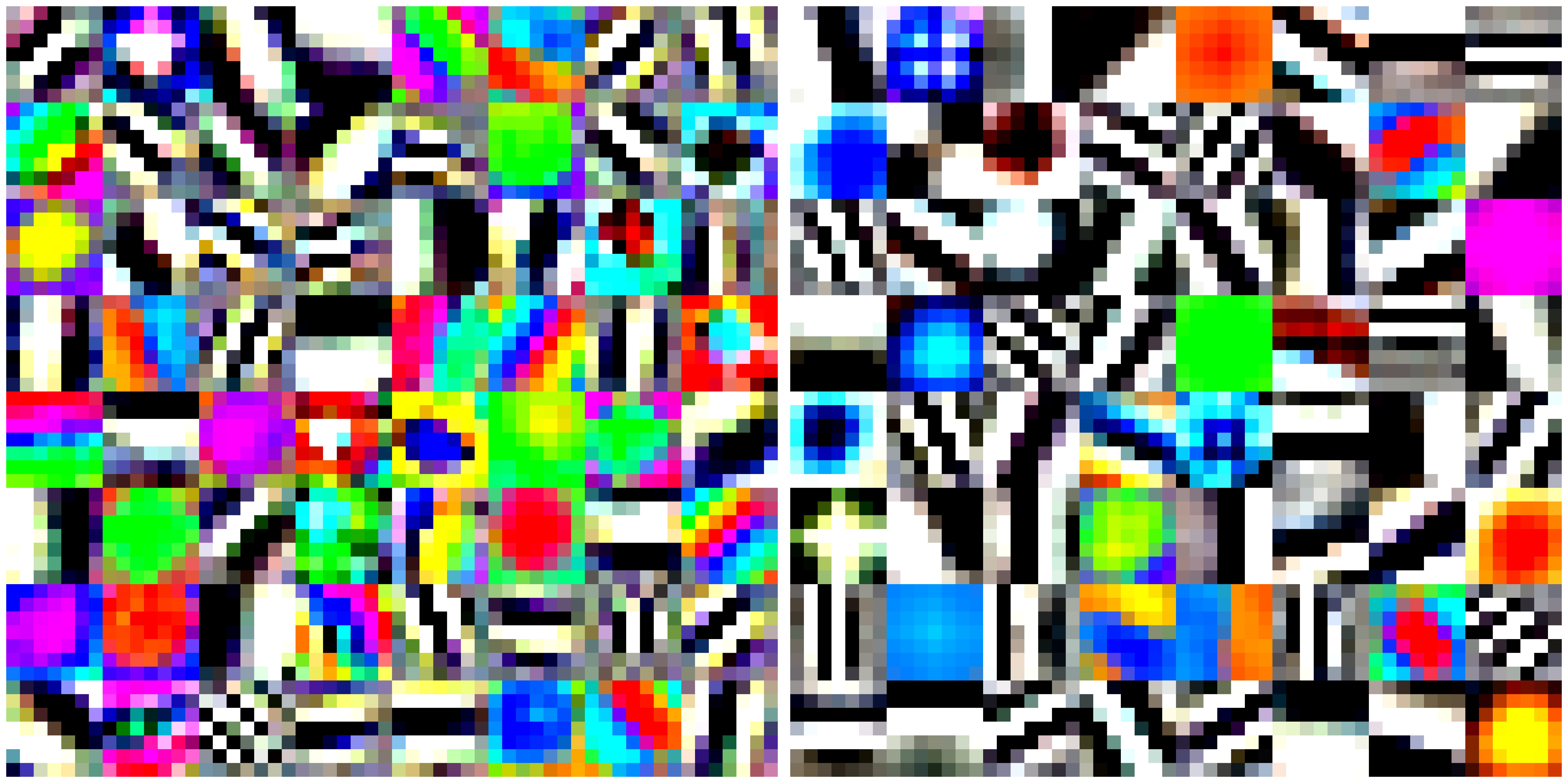}
      {
          \footnotesize
         \begin{flushleft}
            \hspace{1cm}GoogLeNet
            \hspace{1cm}7$\times$7$\times$3
            \hspace{1cm}GoogLeNet-R
         \end{flushleft}
      }

   \end{subfigure}
   \begin{subfigure}{1\linewidth}
      \centering
      \includegraphics[width=1\linewidth]{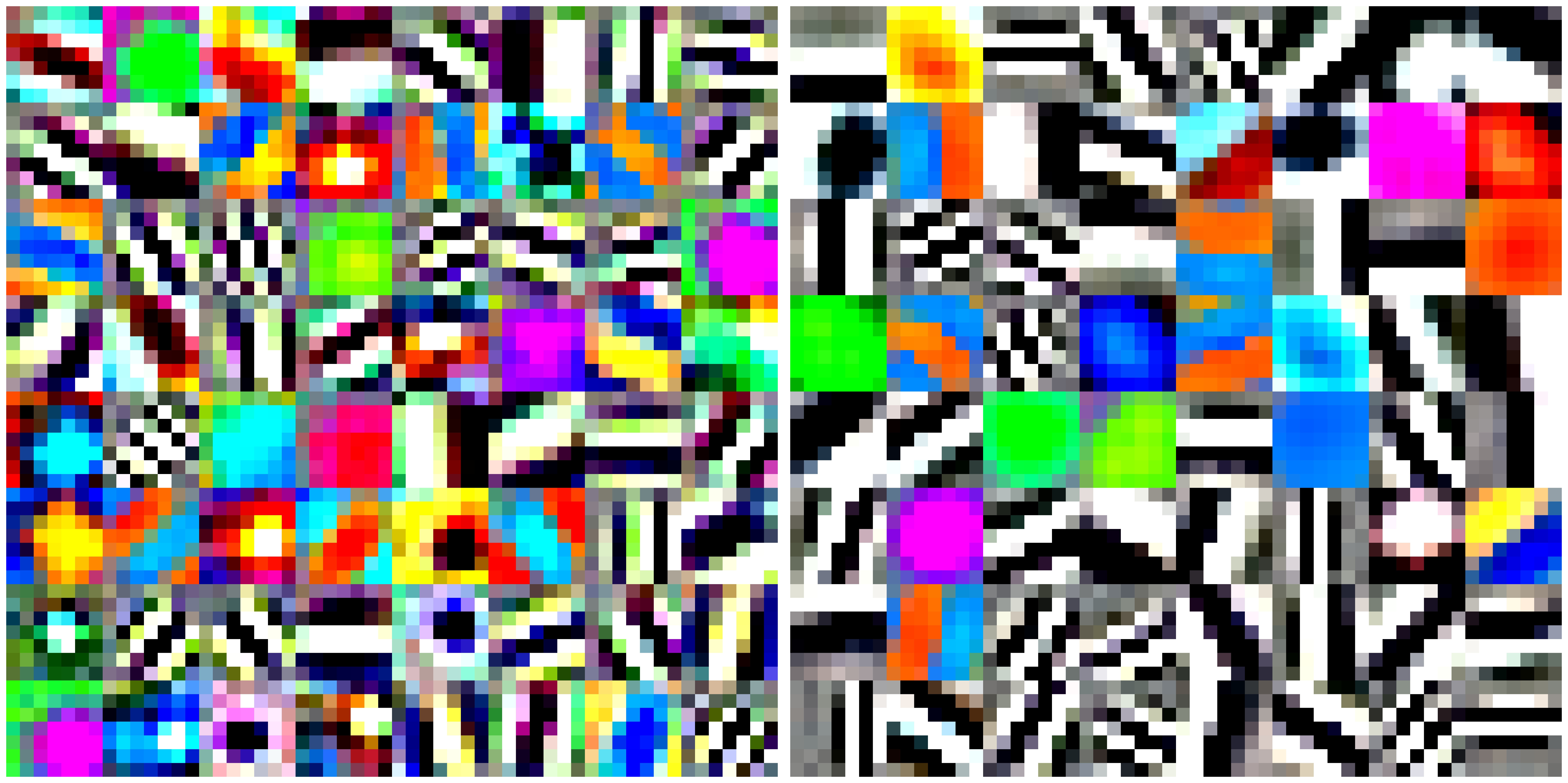}
      {
          \footnotesize
         \begin{flushleft}
            \hspace{1cm}ResNet
            \hspace{1.5cm}7$\times$7$\times$3
            \hspace{1cm}ResNet-R
         \end{flushleft}
      }

   \end{subfigure}
   \caption{
       All 64 \class{conv1} filters of in each standard network (left) and its counterpart (right).
       The filters of R models (right) are smoother and less diverse compared to those in standard models (left).
      Especially, the edge filters of standard networks are noisier and often contain multiple colors in them.
      }
   \label{fig:full_conv1_weights}
\end{figure}

\begin{figure*}[h]
   \centering
   \includegraphics[width=0.8\linewidth]{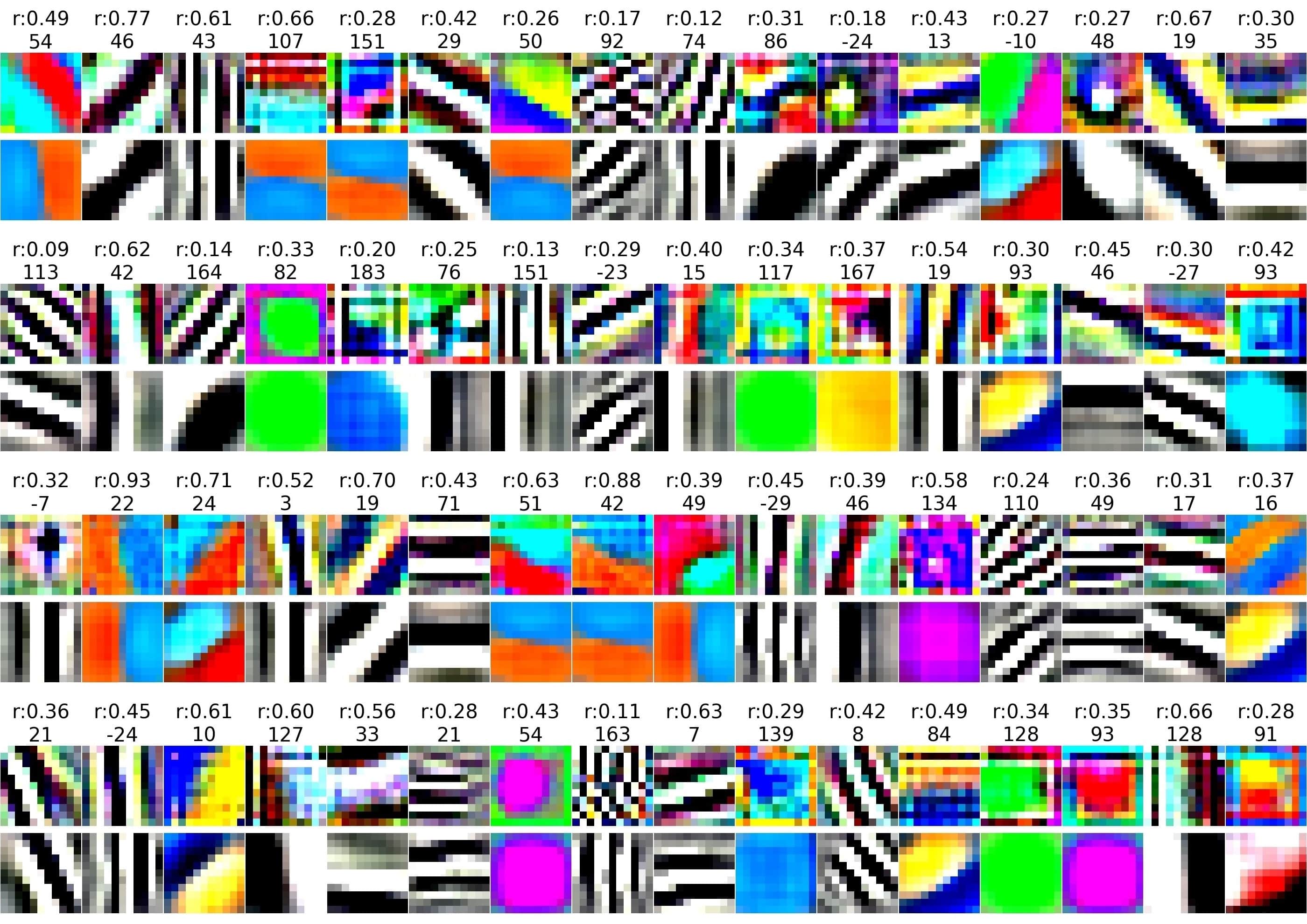}
   \caption{
       \class{conv1} filters of AlexNet-R are smoother than the filters in standard AlexNet.
      In each column, we show an AlexNet filter \class{conv1} filter and their nearest filter (bottom) from the AlexNet-R. 
      Above each pair of filters are their Spearman rank correlation score (\eg \class{r}: 0.36) and their total variation (TV) difference (\ie smoothness differences).
      Standard AlexNet filters are mostly noisier than their nearest R filter (\ie positive TV differences).
%     filter using Spearman Rank correlation.
      }
   \label{fig:compare_corr_alexnet_weights}
\end{figure*}

We visualize the \layer{conv1} weights of the 6 models and plot them side by side to compare their kernel smoothness (Fig.~\ref{fig:full_conv1_weights}). The kernels in R models a consistently smoother than its counter part. For a more straight forward visualization, we use Spearman rank correlation score to pair the similar kernels in AlexNet and AlexNet-R (Fig.~\ref{fig:compare_corr_alexnet_weights}). 

\section{Object and color detectors of AlexNet}
\label{sec:alexnet_object_color}
\begin{figure}[ht]
   \centering
   \begin{subfigure}{1\linewidth}
      \centering
      \includegraphics[width=0.9\linewidth]{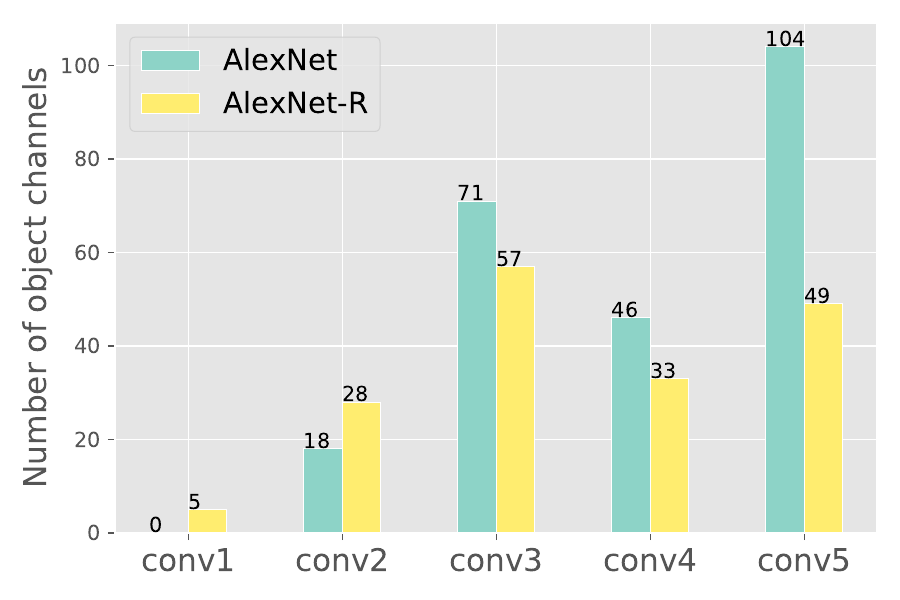}
      \caption{Number of object detectors per AlexNet layer}
%     \vspace{0.2cm}
      \label{fig:Alexnet_object}
   \end{subfigure}   
   \hspace{1mm}
   \begin{subfigure}{1\linewidth}
      \centering
      \includegraphics[width=0.9\linewidth]{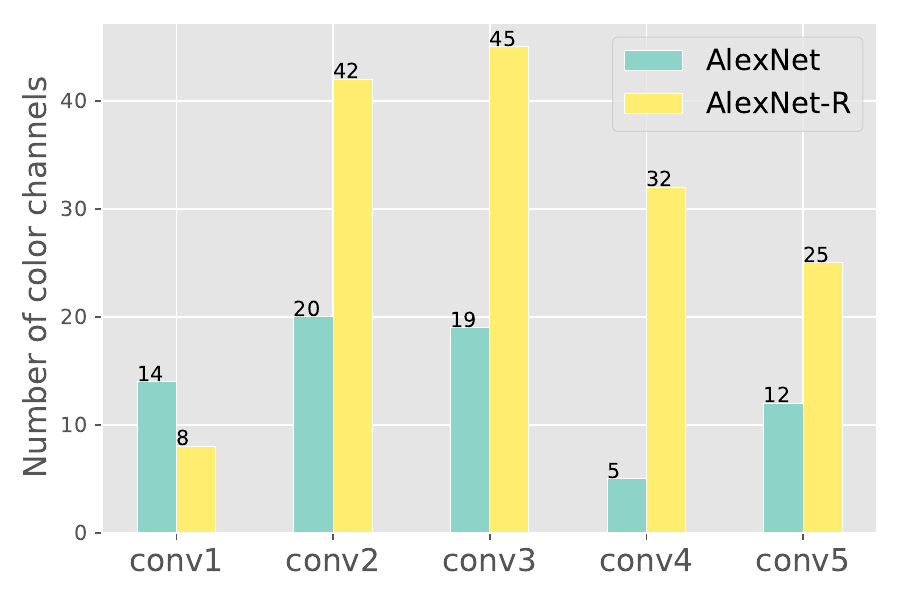}
      \caption{Number of color detectors per AlexNet layer}
%     \vspace{0.2cm}    
      \label{fig:Alexnet_color}
   \end{subfigure}   
   \caption{
       In higher layers (here, \layer{conv4} and \layer{conv5}), AlexNet-R have fewer object detectors but more color detector units compared to standard AlexNet.
       The differences between the two networks increase as we go from lower to higher layers.
       Because both networks share an identical architecture, the plots here demonstrate a substantial shift in the functionality of the neurons as the result of adversarial training---detecting more colors and textures and fewer objects.
       Similar trends were also observed between standard and R models of GoogLeNet and ResNet-50 architectures.
      % The trend is the opposite in very early layers (\layer{conv1}).
   }
   \label{fig:alexnet_object_color}
\end{figure}

We compared the layer-wise difference of \class{object} and \class{color} detectors of AlexNet and AlexNet-R (Fig.~\ref{fig:alexnet_object_color}). We can see in Fig.~\ref{fig:Alexnet_object} that AlexNet has more object detectors and fewer color detectors compared to AlexNet-R.

\section{Total variance (TV) of AlexNet \& AlexNet-R on clean/noisy images}
\label{sec:appendix_alexnet_tv}
\begin{figure}[ht]
   \centering
%  \begin{subfigure}{0.47\linewidth}
%     \centering
%     \includegraphics[width=1.0\linewidth]{images/tv_00.pdf}
%     \caption{
%            \layer{conv1}
%     }
%     \label{fig:si_tv_conv1}
%  \end{subfigure}
   \begin{subfigure}{1\linewidth}
      \centering
      \includegraphics[width=0.8\linewidth]{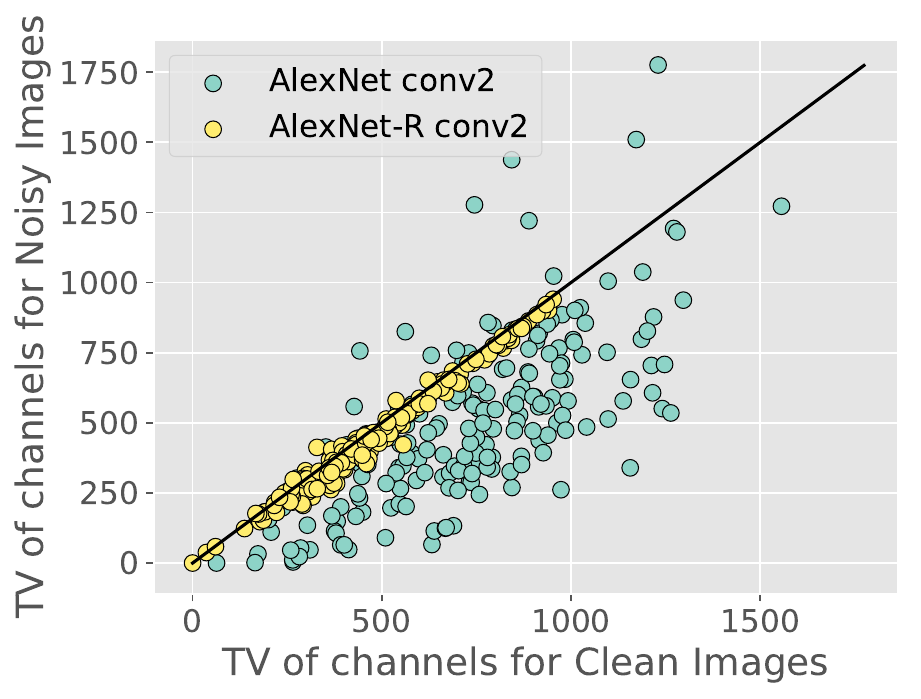}
      \caption{
          \layer{conv2}
      }
      \label{fig:si_tv_conv2}
   \end{subfigure}
%  \begin{subfigure}{0.47\linewidth}
%     \centering
%     \includegraphics[width=1.0\linewidth]{images/tv_02.pdf}
%     \caption{
%            \layer{conv3}
%     }
%     \label{fig:si_tv_conv3}
%  \end{subfigure}
   \begin{subfigure}{0.8\linewidth}
      \centering
      \includegraphics[width=1.0\linewidth]{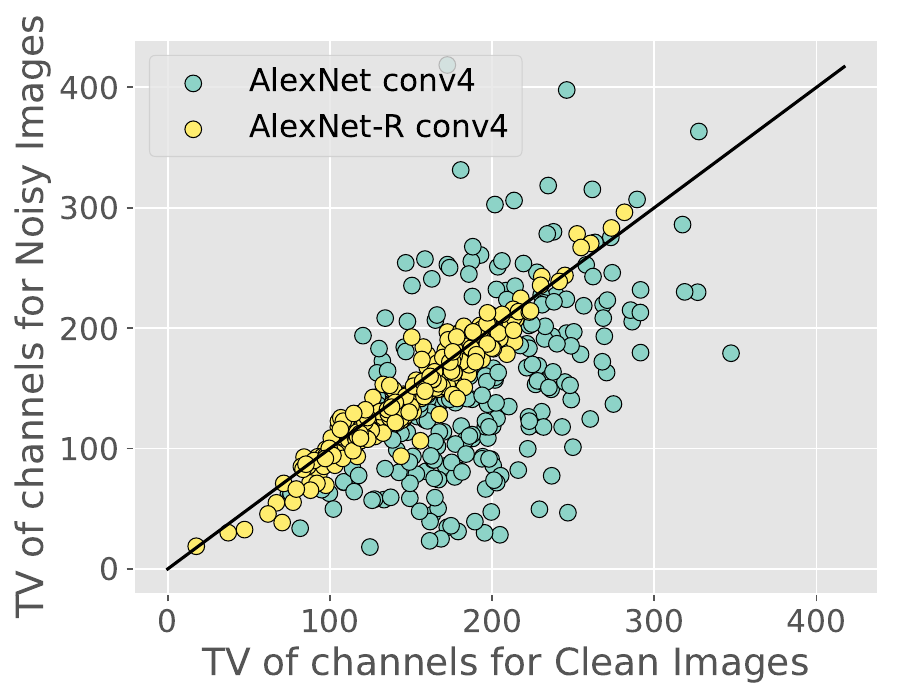}
      \caption{
          \layer{conv4}
      }
      \label{fig:si_tv_conv4}
   \end{subfigure}
   
%  \begin{subfigure}{0.47\linewidth}
%     \centering
%     \includegraphics[width=1.0\linewidth]{images/tv_04.pdf}
%     \caption{
%            \layer{conv5}
%     }
%     \label{fig:si_tv_conv5}
%  \end{subfigure}   
   \caption{
       Each point shows the Total Variation (TV) of the activation maps on clean and noisy images for an AlexNet or AlexNet-R channel.
       We observe a striking difference in \layer{conv1}: The smoothness of R channels remains unchanged before and after noise addition, explaining their superior performance in classifying noisy images.
       While the channel smoothness differences (between two networks) are gradually smaller in higher layers, we still observe R channels are consistently smoother.
      % , changes more in the lower layers of AlexNet.
      % Scatter plot for visualizing the TV of the activation maps of all five \class{conv} layers of AlexNet and AlexNet-R between clean and noisy images.
      % Interestingly, the TV for AlexNet-R does not change much for clean and noisy images thus supporting their superior performances on noisy inputs.
      }
   \label{fig:si_clean_noisy_tv}
\end{figure}

The total variances of AlexNet \& AlexNet-R plot (Fig.~\ref{fig:si_clean_noisy_tv}) shows that the early layer \ie \layer{conv1} in AlexNet-R can filtered out Gaussian noise. The total variances of clean and noisy input results in smilar value in \layer{conv1} in AlexNet-R, this means the noise has been filtered by \layer{conv1}.

\section{ImageNet-C evaluation}
\label{sec:imagenet_c_evaluation}

In Table~\ref{tab:imagenet_c}, we evaluate the validation accuracy of 6 models on 15 common types of image corruptions in ImageNet-C. It turns out that shape biased model does not necessary mean better generalization on these image corruptions. But the R models of AlexNet and ResNet do shows better performance in noise and blur distortions.

\begin{table*}[htbp]
\caption{
    Top-1 accuracy of 6 models (in \%) on all 15 types of image corruptions in ImageNet-C \citep{hendrycks2018benchmarking}.
    On average over all 15 distortion types, R models underperform their standard counterparts.
}
\label{tab:imagenet_c}
\def\arraystretch{1.1}%
\setlength{\tabcolsep}{2pt}
\centering
\begin{tabular}{l|l|r|r|r|r|r|r|r|r}
\hline
 % Peijie's ImageNet-C accuracy
& &  \multicolumn{2}{c|}{AlexNet} & \multicolumn{2}{c|}{GoogLeNet} & \multicolumn{4}{c|}{ResNet} \\ \hline
  &  & Standard & Robust & Standard & Robust & Standard & Robust & AdvProp PGD1 & AdvProp PGD5\\ \hline
\multirow{3}{*}{{(a) Noise}} & {Gaussian} & 11.05 & \textbf{56.10} & \textbf{56.34} & 26.78 & 38.43 & \textbf{45.03} & 49.53 & \textbf{56.39} \\ 
 & {Shot} & 11.15 & \textbf{53.22} & \textbf{50.43} & 24.66 & 35.03 & \textbf{44.07} & 47.56 & \textbf{52.85} \\ 
 & {Impulse} & 8.93 & \textbf{52.49} & \textbf{42.10} & 24.90 & 38.40 & \textbf{41.68} & 49.53 & \textbf{52.39} \\ \hline
\multirow{4}{*}{{(b) Blur}} & {Defocus} & 24.34 & \textbf{28.15} & \textbf{39.63} & 26.52 & \textbf{43.77} & 37.95 & 49.71 & \textbf{56.06} \\ 
 & {Glass} & 22.76 & \textbf{44.50} & 22.62 & \textbf{43.17} & 20.71 & \textbf{51.53} & 30.38 & \textbf{37.74} \\ 
 & {Motion} & 33.87 & \textbf{41.74} & \textbf{44.24} & 41.25 & 44.92 & \textbf{50.16} & 48.27 & \textbf{54.63} \\ 
 & {Zoom} & 38.86 & \textbf{44.27} & 42.58 & \textbf{43.61} & 45.16 & \textbf{52.20} & 49.59 & \textbf{54.97} \\ \hline
\multirow{4}{*}{{(c) Weather}} & {Snow} & 26.59 & \textbf{26.91} & \textbf{52.78} & 15.60 & \textbf{42.13} & 40.39 & 46.01 & \textbf{51.46} \\ 
 & {Frost} & \textbf{21.46} & 13.85 & \textbf{46.14} & 8.38 & \textbf{37.72} & 33.51 & 43.88 & \textbf{50.66} \\ 
 & {Fog} & \textbf{27.86} & 1.64 & \textbf{64.37} & 12.30 & \textbf{56.78} & 3.81 & 54.81 & \textbf{58.64} \\ 
 & {Brightness} & \textbf{77.61} & 62.67 & \textbf{91.60} & 51.14 & \textbf{85.67} & 79.44 & 86.76 & \textbf{88.30} \\ \hline
\multirow{4}{*}{{(d) Digital}} & {Contrast} & \textbf{18.24} & 2.06 & \textbf{78.38} & 22.63 & \textbf{53.67} & 3.46 & 56.57 & \textbf{61.63} \\ 
 & {Elastic} & 75.97 & \textbf{80.98} & \textbf{78.18} & 77.15 & 67.86 & \textbf{82.11} & 74.23 & \textbf{78.44} \\ 
 & {Pixelate} & 57.94 & \textbf{79.46} & \textbf{82.47} & 79.35 & 62.40 & \textbf{83.21} & 67.88 & \textbf{76.38} \\ 
 & {JPEG} & 72.82 & \textbf{85.07} & 80.27 & \textbf{81.72} & 73.66 & \textbf{85.51} & 79.75 & \textbf{82.01} \\ \hline
 
\multirow{4}{*}{{(e) Extra}} & {Speckle Noise} & 17.55 & \textbf{58.42} & \textbf{51.32} & 31.31 & 41.74 & \textbf{52.57} & 52.80 & \textbf{58.07} \\
 & {Gaussian Blur} & 28.68 & 31.26 & \textbf{45.52} & 30.36 & \textbf{47.56} & 41.70 & 55.68 & \textbf{60.43} \\ 
 & {Spatter} & 28.68 & 31.26 & \textbf{45.52} & 30.36 & \textbf{47.56} & 41.70 & 55.68 & \textbf{60.43} \\ 
 & {Saturate} & 46.90 & \textbf{63.66} & \textbf{65.36} & 57.48 & 58.58 & \textbf{70.72} & 66.21 & \textbf{71.44} \\ \hline
\multicolumn{2}{c|}{\textbf{mean Accuracy}} & 37.16 & \textbf{47.34} & \textbf{59.38} & 40.34 & 51.70 & 51.72 & 57.86 & \textbf{62.80} \\ \hline
\end{tabular}
\end{table*}

\section{Examples of shape-less and texture-less images}

\begin{figure*}[h]
   \centering
   {
       \begin{flushleft}
           \hspace{2.5cm}(a) Real
           \hspace{1.2cm}(b) Scrambled
           \hspace{0.8cm}(c) Stylized
           \hspace{1.2cm}(d) B\&W
           \hspace{1.1cm}(e) Silhouette
       \end{flushleft}
   }
   \includegraphics[width=0.8\linewidth]{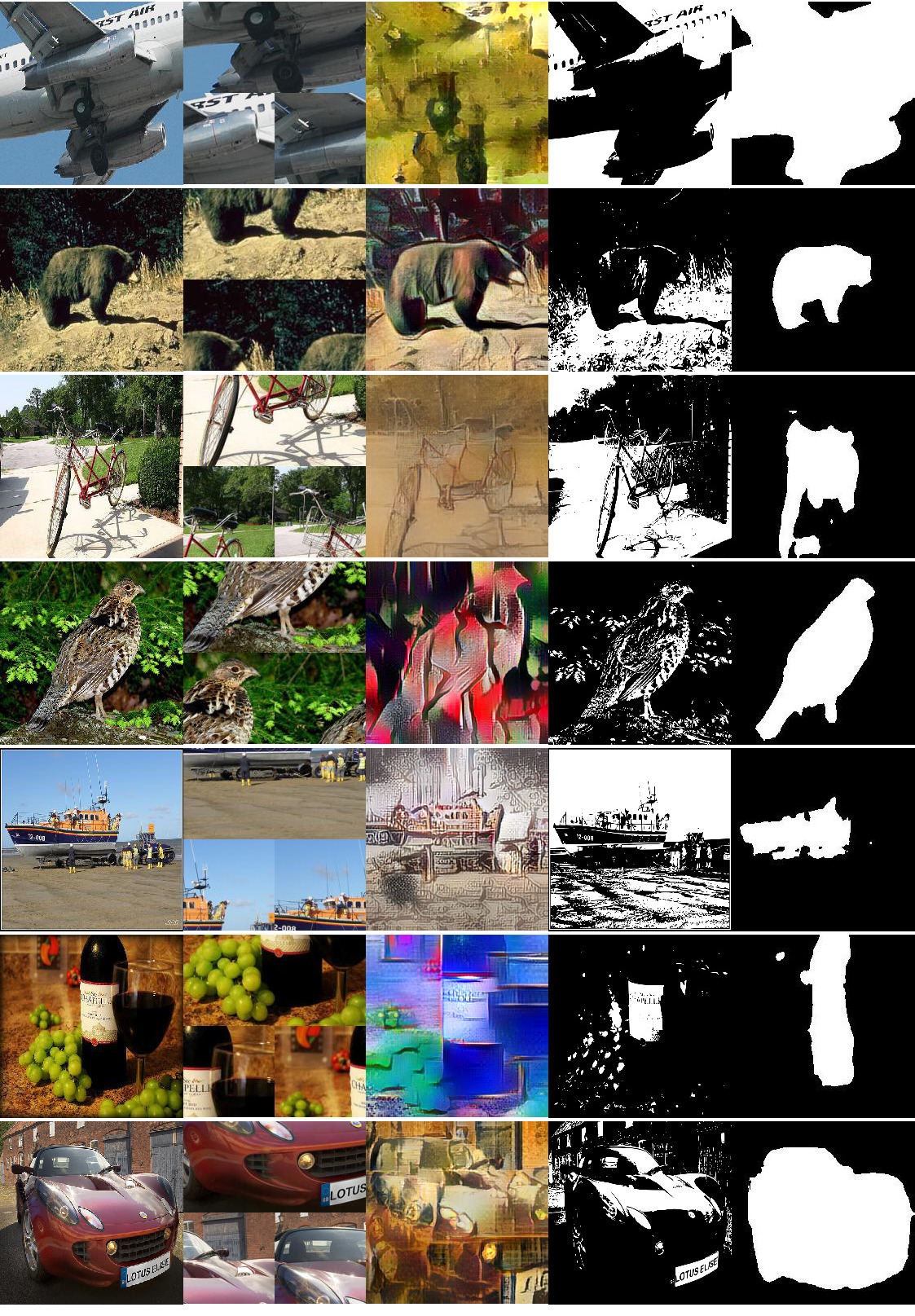}
   \caption{
       Applying different transformation that remove shape/texture on real images. We randomly show an example of 7 out of 16 COCO coarser classes. 
       See Table~\ref{tab:adv_acc} for classification accuracy scores on different images distortion dataset in 1000 classes(Except for Silhouette). \textit{*Note: Silhouette are validate in 16 COCO coarse classes.}
      % These images were used to compute the accuracy in the stylized column of Table~\ref{tab:adv_acc}.
   }
   \label{fig:shape_texture-less}
\end{figure*}

We randomly choose one image in 7 COCO coarser classes (out of 16) and plot the corresponding shape-less and texture-less version (Fig.~\ref{fig:shape_texture-less}).

\section{Visualizing channel preference via cue-conflict and NetDissect}
\label{sec:alexnet_channel_preference}
\begin{figure*}[ht]
   \centering
   \includegraphics[width=1\linewidth]{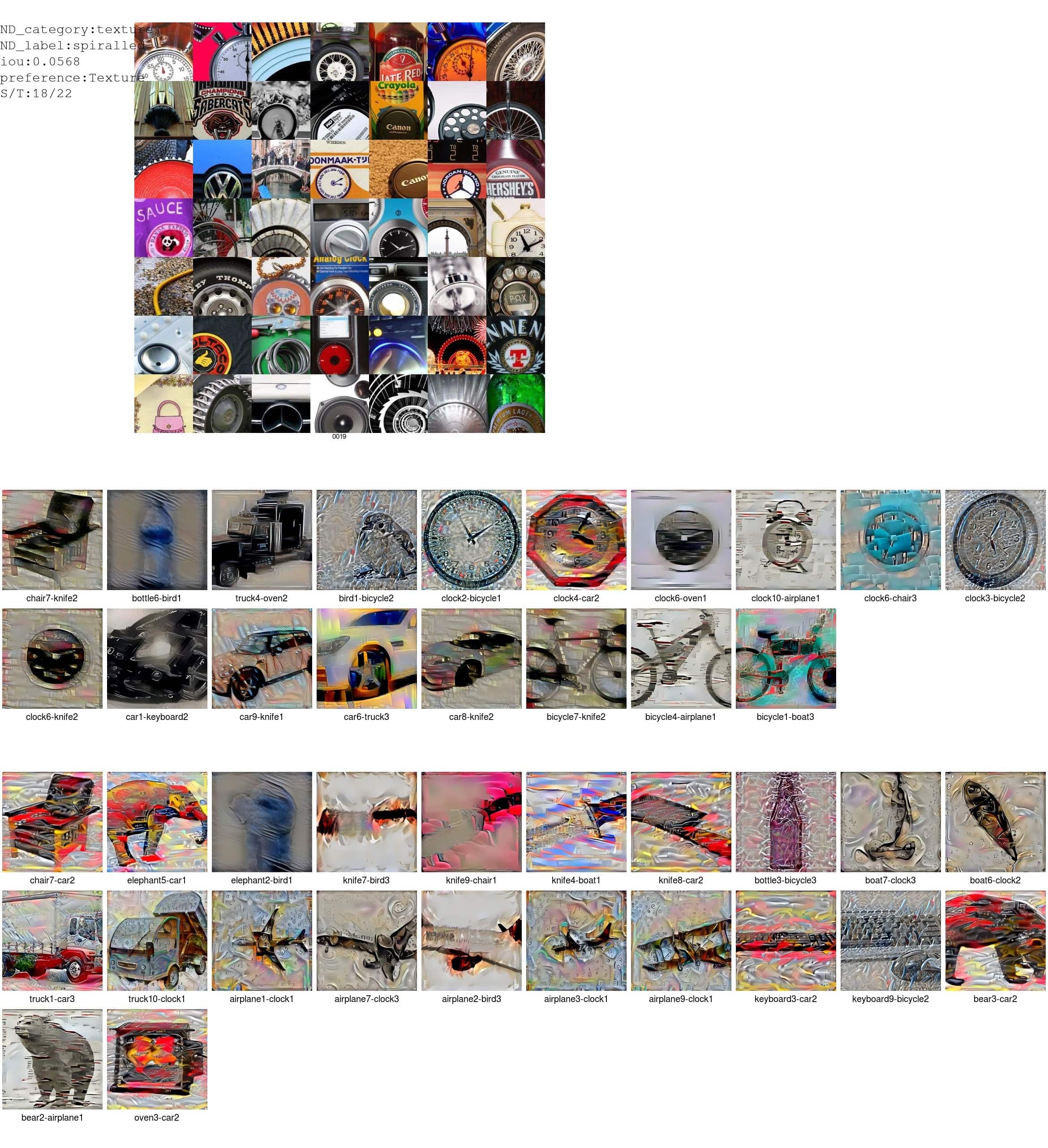}
   \caption{ 
   AlexNet \layerunit{conv4}{19} with Shape and Texture scores of 18 and 22, respectively. 
   It has a NetDissect label of \class{spiralled} (IoU: 0.0568) under \class{texture} category.
   Although this neuron is in NetDissect \class{texture} category, the misclassified images suggest that this neuron helps in both shape- and texture-based recognition. \textbf{Top:} Top-49 images that highest-activated this channel. \textbf{Middle:} Mis-classified images in \class{shape} category (18 images). \textbf{Bottom:} Mis-classified images in \class{texture} category (22 images).
      } 
   \label{fig:Appendix_top49_cue_conflict_visulization_1}
\end{figure*}

\begin{figure*}[ht]
   \centering
   \includegraphics[width=1\linewidth]{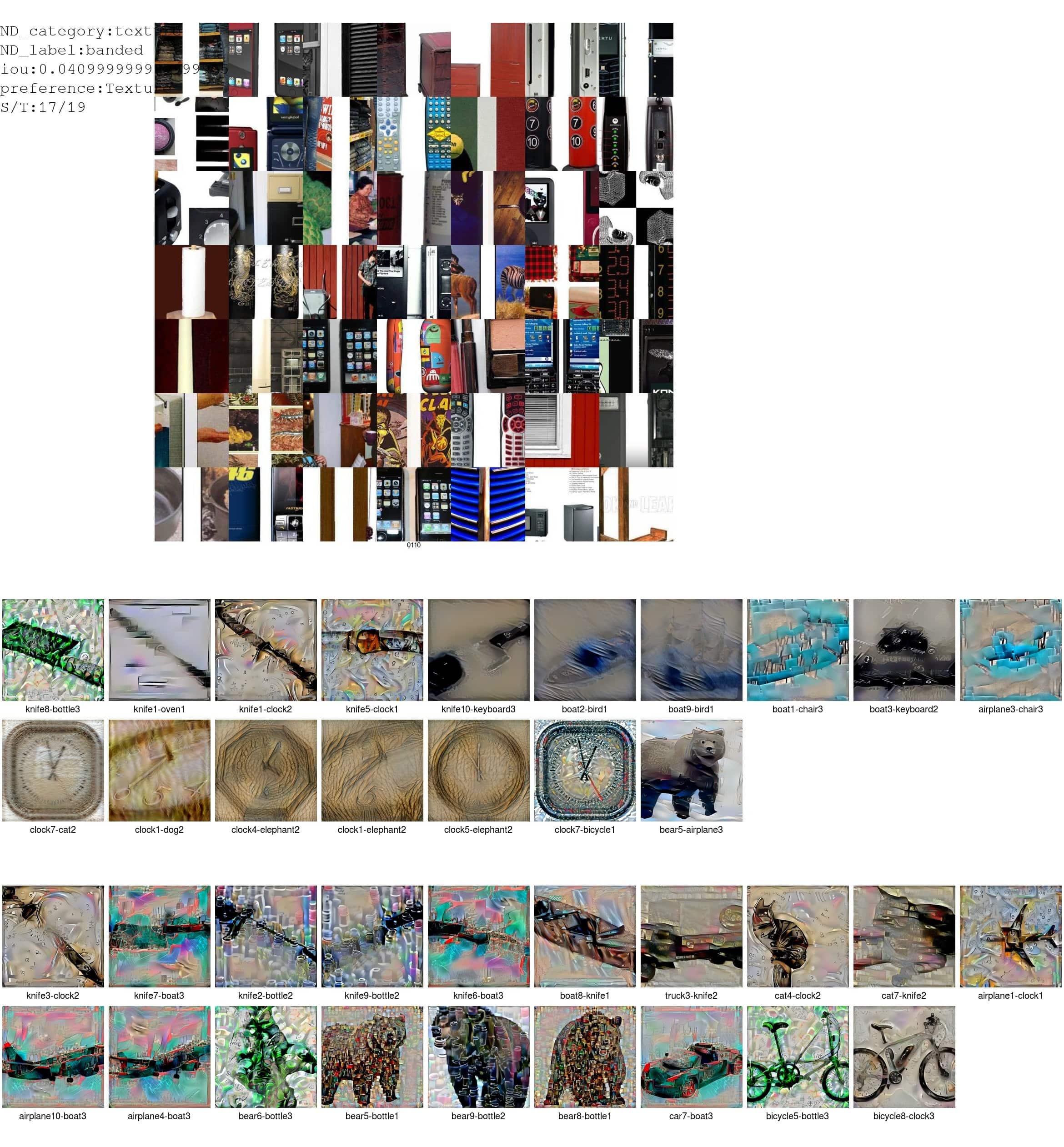}
   \caption{AlexNet-R \layerunit{conv5}{110} with Shape and Texture scores of 17 and 19, respectively. It has a NetDissect label of \class{banded} (IoU: 0.0409) under \class{texture} category. 
   This neuron has almost equal Shape and Texture scores and  is useful in detecting both the shape and textures of knives and bottles at the same time. 
   \textbf{Top:} Top-49 images that highest-activated this channel. \textbf{Middle:} Mis-classified images in \class{shape} category. \textbf{Bottom:} Mis-classified images in \class{texture} category.
      } 
   \label{fig:Appendix_top49_cue_conflict_visulization_7}
\end{figure*}

\begin{figure*}[ht]
   \centering
   \includegraphics[width=1\linewidth]{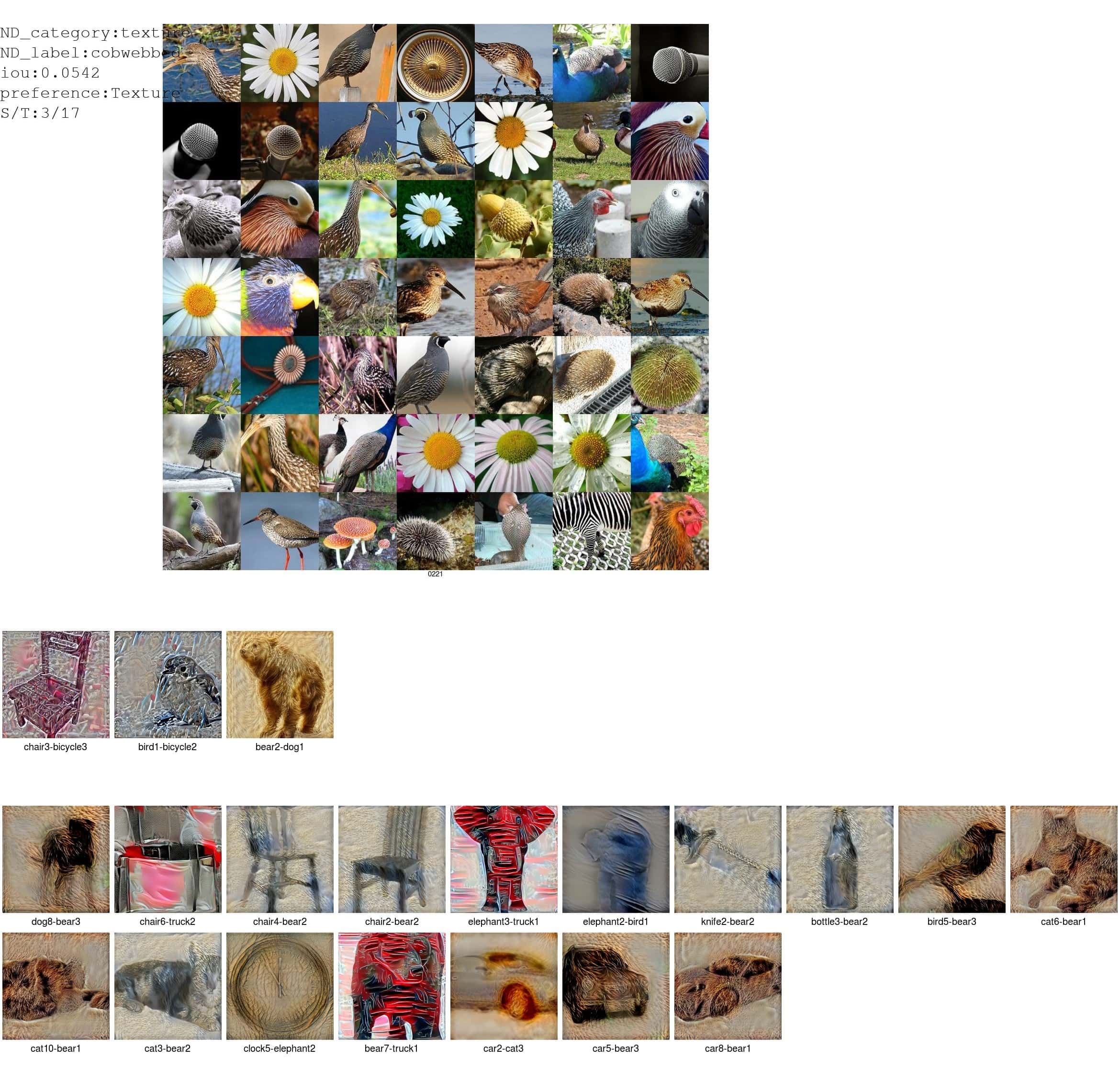}
   \caption{  AlexNet \layerunit{conv5}{221} with Shape and Texture scores of 3 and 17, respectively. It has a NetDissect label of \class{cobwebbed} (IoU: 0.0542) under \class{texture} category.  
   This is a heavily texture-biased neuron that helps networks detect animals by their fur textures. \textbf{Top:} Top-49 images that highest-activated this channel. \textbf{Middle:} Mis-classified images in \class{shape} category. \textbf{Bottom:} Mis-classified images in \class{texture} category.
      } 
   \label{fig:Appendix_top49_cue_conflict_visulization_4}
\end{figure*}

In Fig.~\ref{fig:Appendix_top49_cue_conflict_visulization_1}-~\ref{fig:Appendix_top49_cue_conflict_visulization_4}, we show three samples of the channel preferences experiment. In each of the sample, \textbf{Top} is the top-49 images of the channels (Similar to Fig.~\ref{fig:top49_striped}). On the \textbf{Middle \& Bottom}, we zero-out the corresponding channel and re-run the conflict test to find out the images that were mis-classified. \ie Fig.\ref{fig:Appendix_top49_cue_conflict_visulization_1} the clock images in \textbf{Middle} were classified into shape category by cue-conflict test. After zeroing-out the channel, the network lose the ability to classify the image into shape category.

% top49 images visualization

\begin{figure}[htbp]
   \centering
   \begin{subfigure}{1\linewidth}
   \includegraphics[width=1.0\linewidth]{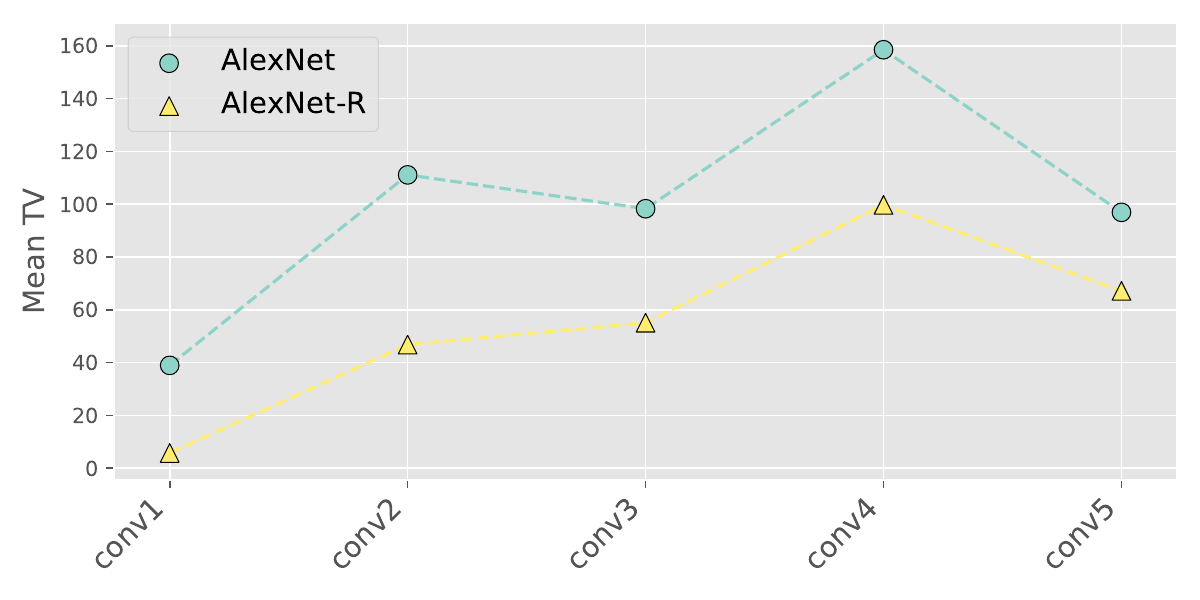}
   \caption{Mean layer-wise kernel TV of AlexNet and AlexNet-R  
       }
   \label{fig:alexnet_kernel_tv}
   \end{subfigure}
      \begin{subfigure}{1\linewidth}
   \includegraphics[width=1.0\linewidth]{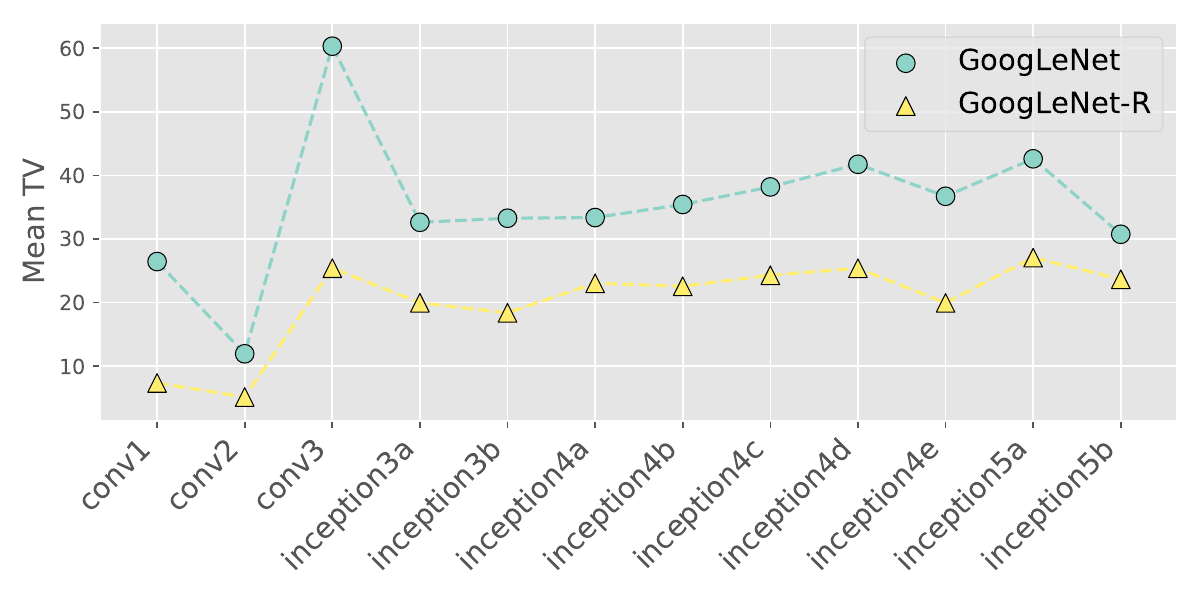}
   \caption{Mean layer-wise kernel TV of GoogLeNet and GoogLeNet-R 
      }
   \label{fig:googlenet_kernel_tv}
   \end{subfigure}
   \begin{subfigure}{1\linewidth}
   \includegraphics[width=1\linewidth]{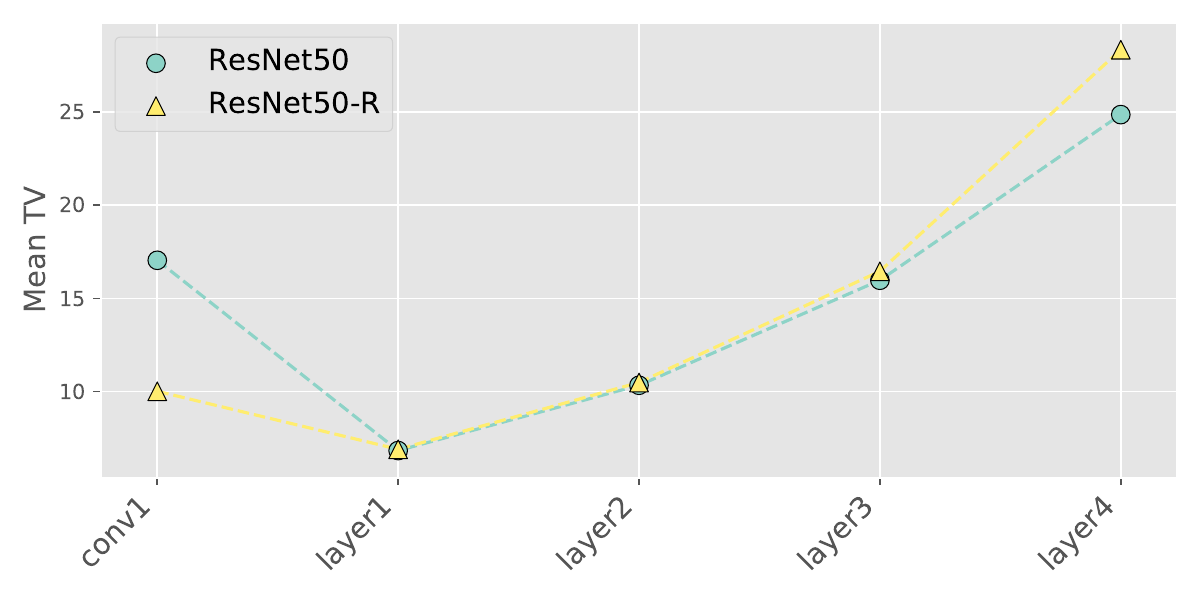}
   \label{fig:resnet_kernel_tv}
   \caption{Mean layer-wise kernel TV of ResNet and ResNet-R}
   \end{subfigure}
   \caption{
   For all main \layer{conv} layers, AlexNet-R filters are smoother (\ie lower TV mean) than their counterparts in AlexNet (a).
   The same observation was found in GoogLeNet-R vs. GoogLeNet comparison (b).
   In ResNet-R, its early filters at \layer{conv1} are also smoother than those in ResNet.
%  In early convolutional layers, the kernels are much smoother in R networks. i.e. (a)\&(b) the kernel is smoother in first few \layer{conv} layers. (c) is a bit special since the \layer{layer*} is residual blocks, but we can still see that \layer{conv1} in robust network has much smoother kernel compared to its' counter part. 
%  Note: For ResNet architectures, we computed the mean TV of all \layer{conv} layers inside each residual (inception) blocks. \anh{For GoogLeNet or ResNet??} See appendix Sec.~\ref{sec:list_netdissect_layers} for details.
   }
   \label{fig:Appendix_kernel_tv}
\end{figure}

\begin{figure*}[htbp]
   \centering
   \begin{subfigure}{1\linewidth}
   \centering
   \includegraphics[width=1\linewidth]{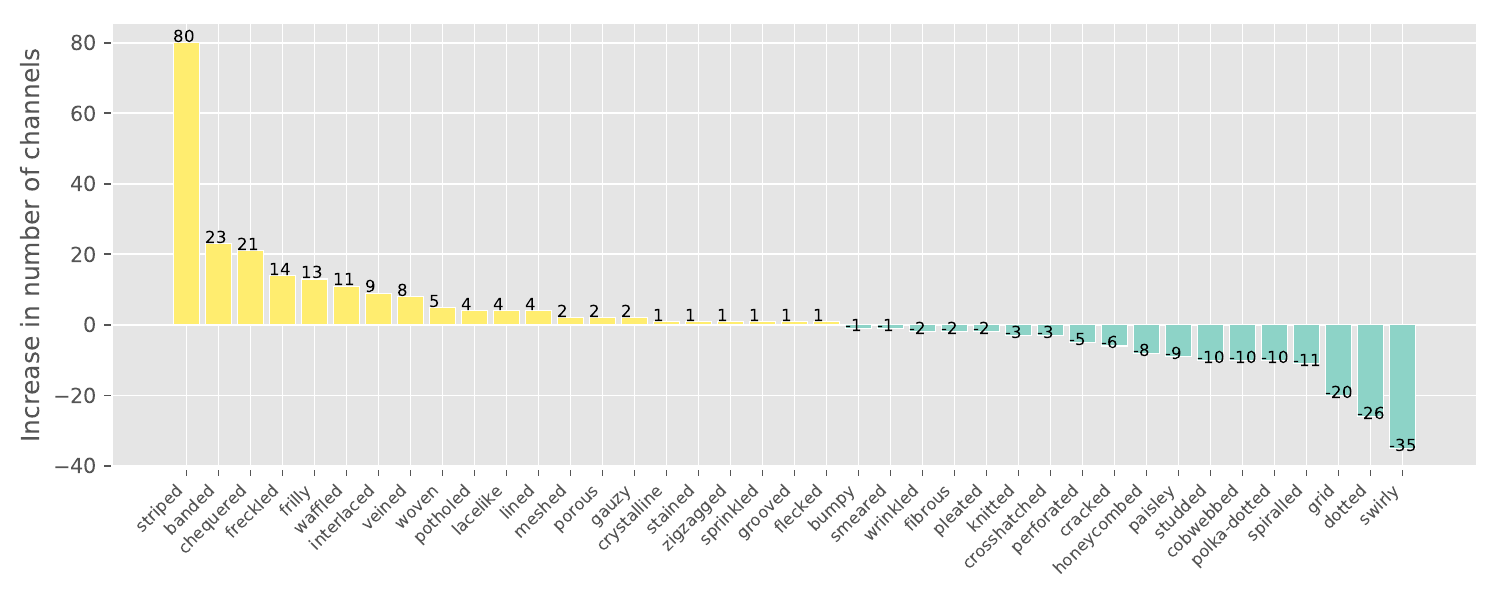}
   \caption{
      Differences in \class{texture} channels between AlexNet and AlexNet-R}
   \label{fig:alexnet_texture_diff}
   \end{subfigure}
   
   \begin{subfigure}{1\linewidth}
      \centering
   \includegraphics[width=1.0\linewidth]{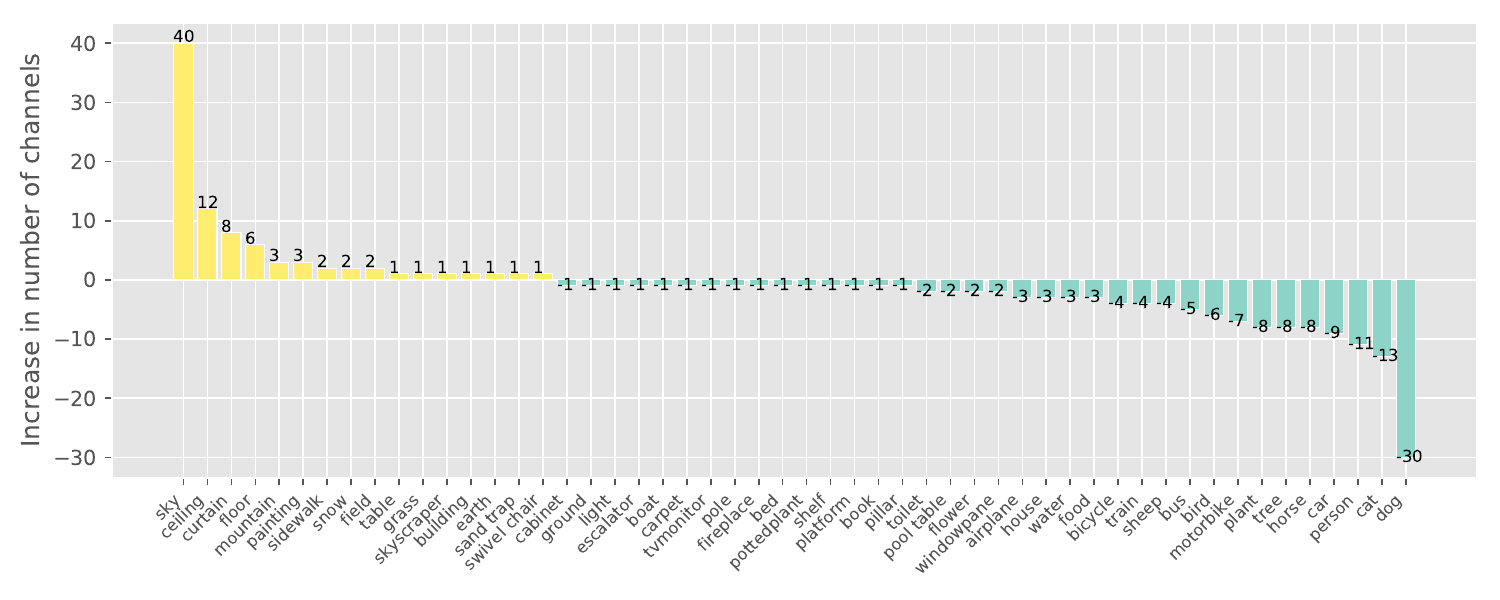}
   \caption{
      Differences in \class{object} channels between AlexNet and AlexNet-R}
   \label{fig:alexnet_object_diff}
   \end{subfigure}
   \caption{
      % \todo{y-axis labels, for a \& b are too big}
       In each bar plot, we column shows the difference in the number of channels (between AlexNet-R and AlexNet) for a given concept \eg \class{striped} or \class{banded}.
       That is, yellow bars (\ie positive numbers) show the count of channels that the R model has more than the standard network in the same concept.
       Vice versa, teal bars represent the concepts that R models have fewer channels.
       The NetDissect concept names are given in the x-axis.
       \textbf{Top:} In the \class{texture} category, the R model has a lot more simple texture patterns \eg \class{striped} and \class{banded} (see Fig.~\ref{fig:alexnet_top5_concept} for example patterns in these concepts).
       \textbf{Bottom:} In the \class{object} category, AlexNet-R often prefers simpler-object detectors \eg \class{sky} or \class{ceiling} (Fig.~\ref{fig:alexnet_object_diff}; leftmost) while the standard network has more complex-objects detectors \eg \class{dog} and \class{cat} (Fig.~\ref{fig:alexnet_object_diff}; rightmost).
        }   
\end{figure*}

\begin{figure*}[ht]
    \centering
    \includegraphics[width=1.0\linewidth]{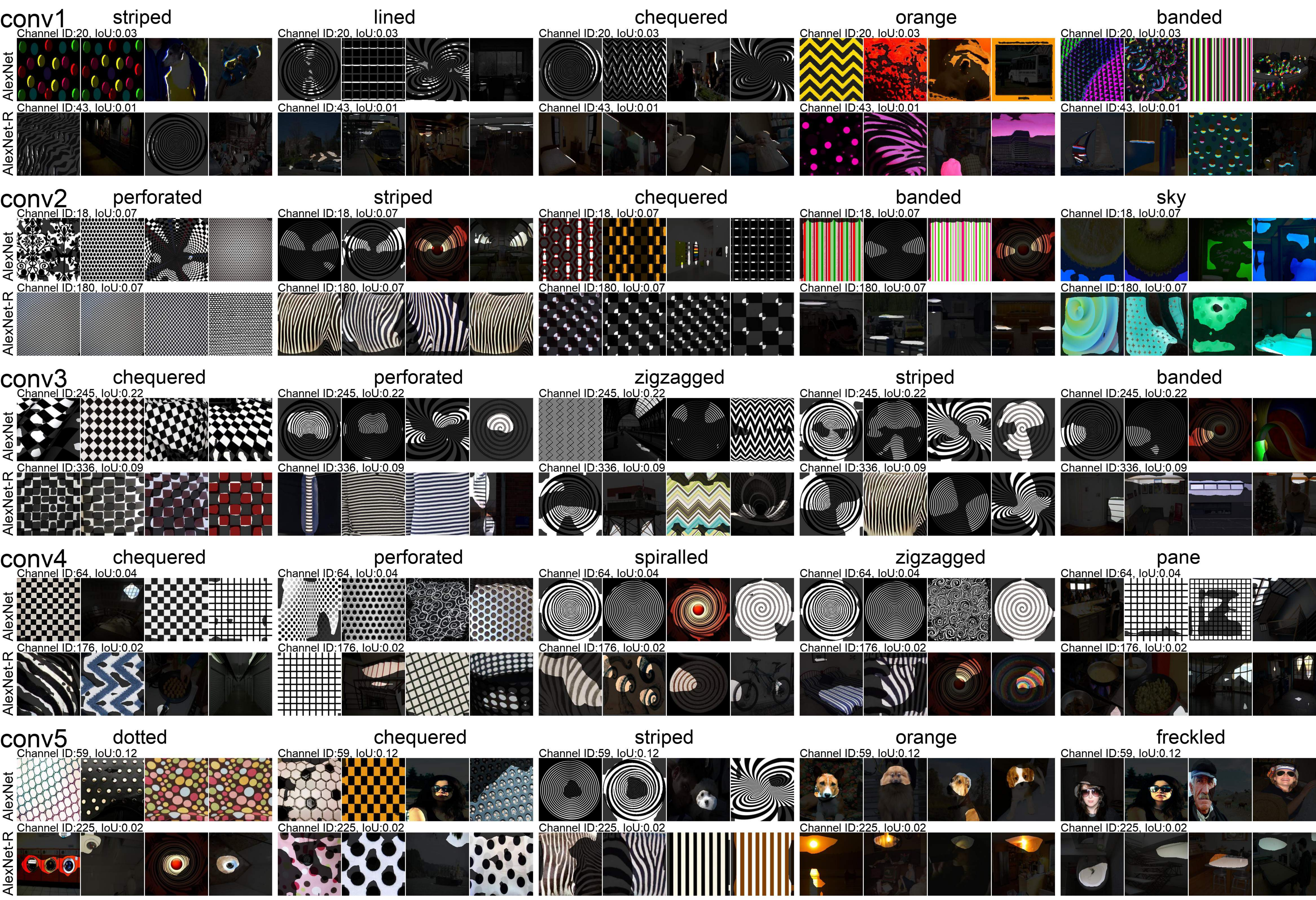}
    \caption{
        The NetDissect images preferred by the channels in the top-5 most important concepts in AlexNet (\ie highest accuracy drop when zeroed out; see Sec.~\ref{sec:netdissect_robust_difference}).
        For each concept, we show the highest-IoU channels.
        % NetDissect result of top 5 concepts in AlexNet (accuracy drop of on Clean ImageNet). These images are from Broden dataset in NetDissect, where the corresponding channel has highest activation.
    }
    \label{fig:alexnet_top5_concept}
\end{figure*}

\begin{figure*}[htbp]
   \centering
   \begin{subfigure}{0.49\linewidth}
   \centering
   \includegraphics[width=0.8\linewidth]{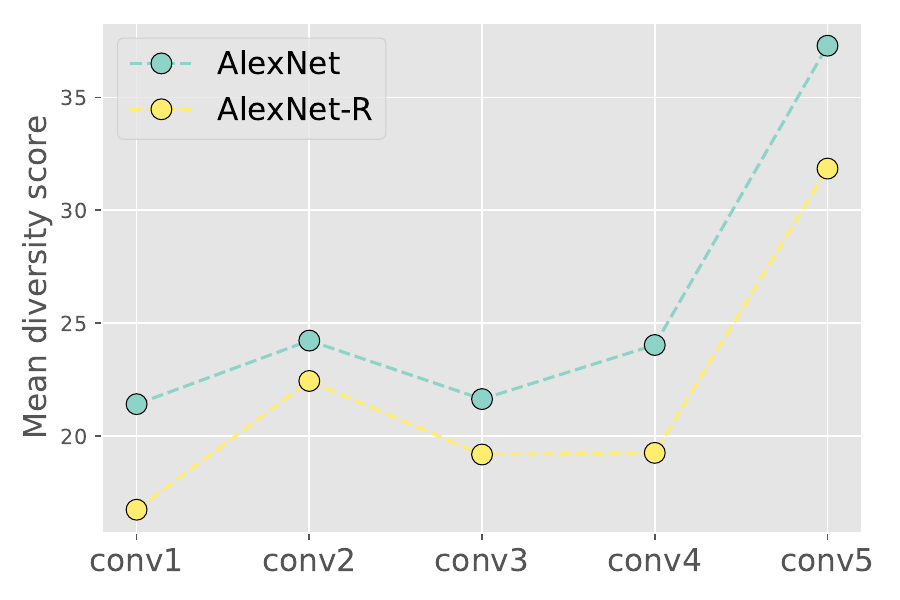}
   \caption{
      AlexNet layer-wise mean diversity}
   \label{fig:alexnet_diversity_plot}
   \end{subfigure}
   \begin{subfigure}{0.49\linewidth}
      \centering
   \includegraphics[width=0.8\linewidth]{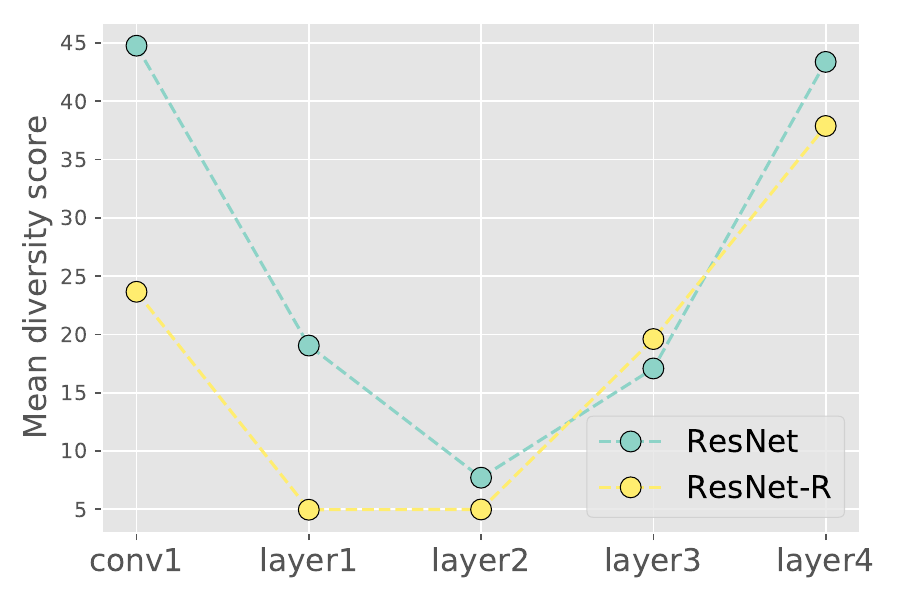}
   \caption{
      ResNet layer-wise mean diversity}
   \label{fig:resnet_diversity_plot}
   \end{subfigure}
   
   \caption{
       In each plot, we show the mean diversity scores across all channels in each layer.
      % Diversity plot of AlexNet and ResNet. 
       Both AlexNet-R and ResNet-R consistently have channels with lower diversity scores (\ie detecting fewer unique concepts) than the standard counterparts.
      % The diversity scores are lower in both R networks. The lower score in diversity matches the smoother filter finding in Sec.\ref{sec:smooth_tv}
       }
   \label{fig:diversity_plot}
\end{figure*}

\clearpage
% \bibliographystyle{cas-model2-names}
\bibliographystyle{apalike}

\bibliography{references}

\begin{thebibliography}{}

\bibitem[Arpit et~al., 2017]{arpit2017closer}
Arpit, D., Jastrz{{e}}bski, S., Ballas, N., Krueger, D., Bengio, E., Kanwal,
  M.~S., Maharaj, T., Fischer, A., Courville, A., Bengio, Y., et~al. (2017).
\newblock A closer look at memorization in deep networks.
\newblock In {\em Proceedings of the 34th International Conference on Machine
  Learning-Volume 70}, pages 233--242. JMLR. org.

\bibitem[Bansal et~al., 2020]{bansal2020sam}
Bansal, N., Agarwal, C., and Nguyen, A. (2020).
\newblock Sam: The sensitivity of attribution methods to hyperparameters.
\newblock In {\em Proceedings of the ieee/cvf conference on computer vision and
  pattern recognition}, pages 8673--8683.

\bibitem[Bau et~al., 2017]{bau2017network}
Bau, D., Zhou, B., Khosla, A., Oliva, A., and Torralba, A. (2017).
\newblock Network dissection: Quantifying interpretability of deep visual
  representations.
\newblock In {\em Proceedings of the IEEE conference on computer vision and
  pattern recognition}, pages 6541--6549.

\bibitem[Brendel and Bethge, 2019]{brendel2018bagnets}
Brendel, W. and Bethge, M. (2019).
\newblock Approximating cnns with bag-of-local-features models works
  surprisingly well on imagenet.
\newblock {\em International Conference on Learning Representations}.

\bibitem[Caesar et~al., 2018]{caesar2018coco}
Caesar, H., Uijlings, J., and Ferrari, V. (2018).
\newblock Coco-stuff: Thing and stuff classes in context.
\newblock In {\em Proceedings of the IEEE Conference on Computer Vision and
  Pattern Recognition}, pages 1209--1218.

\bibitem[Chen et~al., 2017]{chen2017deeplab}
Chen, L.-C., Papandreou, G., Kokkinos, I., Murphy, K., and Yuille, A.~L.
  (2017).
\newblock Deeplab: Semantic image segmentation with deep convolutional nets,
  atrous convolution, and fully connected crfs.
\newblock {\em IEEE transactions on pattern analysis and machine intelligence},
  40(4):834--848.

\bibitem[Cichy et~al., 2016]{cichy2016comparison}
Cichy, R.~M., Khosla, A., Pantazis, D., Torralba, A., and Oliva, A. (2016).
\newblock Comparison of deep neural networks to spatio-temporal cortical
  dynamics of human visual object recognition reveals hierarchical
  correspondence.
\newblock {\em Scientific reports}, 6(1):1--13.

\bibitem[De~Palma et~al., 2019]{de2019random}
De~Palma, G., Kiani, B., and Lloyd, S. (2019).
\newblock Random deep neural networks are biased towards simple functions.
\newblock In {\em Advances in Neural Information Processing Systems}, pages
  1962--1974.

\bibitem[Engstrom et~al., 2019]{robustness}
Engstrom, L., Ilyas, A., Santurkar, S., and Tsipras, D. (2019).
\newblock Robustness (python library).

\bibitem[Engstrom et~al., 2020]{engstrom2020adversarial}
Engstrom, L., Ilyas, A., Santurkar, S., Tsipras, D., Tran, B., and Madry, A.
  (2020).
\newblock Adversarial robustness as a prior for learned representations.

\bibitem[Ford et~al., 2019]{ford2019adversarial}
Ford, N., Gilmer, J., and Cubuk, E.~D. (2019).
\newblock Adversarial examples are a natural consequence of test error in
  noise.

\bibitem[Gatys et~al., 2016]{gatys2016image}
Gatys, L.~A., Ecker, A.~S., and Bethge, M. (2016).
\newblock Image style transfer using convolutional neural networks.
\newblock In {\em Proceedings of the IEEE conference on computer vision and
  pattern recognition}, pages 2414--2423.

\bibitem[Geirhos et~al., 2019]{geirhos2018imagenet}
Geirhos, R., Rubisch, P., Michaelis, C., Bethge, M., Wichmann, F.~A., and
  Brendel, W. (2019).
\newblock Imagenet-trained {CNN}s are biased towards texture; increasing shape
  bias improves accuracy and robustness.
\newblock In {\em International Conference on Learning Representations}.

\bibitem[Geirhos et~al., 2018]{geirhos2018generalisation}
Geirhos, R., Temme, C.~R., Rauber, J., Sch{\"u}tt, H.~H., Bethge, M., and
  Wichmann, F.~A. (2018).
\newblock Generalisation in humans and deep neural networks.
\newblock In {\em Advances in Neural Information Processing Systems}, pages
  7538--7550.

\bibitem[Gilmer et~al., 2019]{gilmer2019adversarial}
Gilmer, J., Ford, N., Carlini, N., and Cubuk, E. (2019).
\newblock Adversarial examples are a natural consequence of test error in
  noise.
\newblock In {\em International Conference on Machine Learning}, pages
  2280--2289.

\bibitem[He et~al., 2016]{he2016deep}
He, K., Zhang, X., Ren, S., and Sun, J. (2016).
\newblock Deep residual learning for image recognition.
\newblock In {\em Proceedings of the IEEE conference on computer vision and
  pattern recognition}, pages 770--778.

\bibitem[Hendrycks and Dietterich, 2019]{hendrycks2018benchmarking}
Hendrycks, D. and Dietterich, T. (2019).
\newblock Benchmarking neural network robustness to common corruptions and
  perturbations.
\newblock In {\em International Conference on Learning Representations}.

\bibitem[{ImageMagick}, 2020]{imagemagick}
{ImageMagick} (2020).
\newblock Imagemagick.

\bibitem[Krizhevsky et~al., 2012]{krizhevsky2012imagenet}
Krizhevsky, A., Sutskever, I., and Hinton, G.~E. (2012).
\newblock Imagenet classification with deep convolutional neural networks.
\newblock In {\em Advances in neural information processing systems}, pages
  1097--1105.

\bibitem[Madry et~al., 2018]{madry2017towards}
Madry, A., Makelov, A., Schmidt, L., Tsipras, D., and Vladu, A. (2018).
\newblock Towards deep learning models resistant to adversarial attacks.
\newblock In {\em International Conference on Learning Representations}.

\bibitem[Nguyen et~al., 2016a]{nguyen2016synthesizing}
Nguyen, A., Dosovitskiy, A., Yosinski, J., Brox, T., and Clune, J. (2016a).
\newblock Synthesizing the preferred inputs for neurons in neural networks via
  deep generator networks.
\newblock In {\em Advances in neural information processing systems}, pages
  3387--3395.

\bibitem[Nguyen et~al., 2016b]{nguyen2016multifaceted}
Nguyen, A., Yosinski, J., and Clune, J. (2016b).
\newblock Multifaceted feature visualization: Uncovering the different types of
  features learned by each neuron in deep neural networks.
\newblock In {\em Visualization for Deep Learning Workshop, ICML conference}.

\bibitem[Nguyen et~al., 2019]{nguyen2019understanding}
Nguyen, A., Yosinski, J., and Clune, J. (2019).
\newblock Understanding neural networks via feature visualization: A survey.
\newblock In {\em Explainable AI: Interpreting, Explaining and Visualizing Deep
  Learning}, pages 55--76. Springer.

\bibitem[PyTorch, 2019]{torchvis88:online}
PyTorch (2019).
\newblock torchvision.models — pytorch master documentation.
\newblock \url{https://pytorch.org/docs/stable/torchvision/models.html}.
\newblock (Accessed on 09/21/2019).

\bibitem[Rahaman et~al., 2019]{rahaman2019on}
Rahaman, N., Baratin, A., Arpit, D., Dräxler, F., Lin, M., Hamprecht, F.~A.,
  Bengio, Y., and Courville, A.~C. (2019).
\newblock On the spectral bias of neural networks.
\newblock In {\em ICML}, pages 5301--5310.

\bibitem[Rajalingham et~al., 2018]{rajalingham2018large}
Rajalingham, R., Issa, E.~B., Bashivan, P., Kar, K., Schmidt, K., and DiCarlo,
  J.~J. (2018).
\newblock Large-scale, high-resolution comparison of the core visual object
  recognition behavior of humans, monkeys, and state-of-the-art deep artificial
  neural networks.
\newblock {\em Journal of Neuroscience}, 38(33):7255--7269.

\bibitem[Rozsa and Boult, 2019]{rozsa2019improved}
Rozsa, A. and Boult, T.~E. (2019).
\newblock Improved adversarial robustness by reducing open space risk via tent
  activations.
\newblock {\em arXiv preprint arXiv:1908.02435}.

\bibitem[Rudin et~al., 1992]{rudin1992nonlinear}
Rudin, L.~I., Osher, S., and Fatemi, E. (1992).
\newblock Nonlinear total variation based noise removal algorithms.
\newblock {\em Physica D: nonlinear phenomena}, 60(1-4):259--268.

\bibitem[Russakovsky et~al., 2015]{russakovsky2015imagenet}
Russakovsky, O., Deng, J., Su, H., Krause, J., Satheesh, S., Ma, S., Huang, Z.,
  Karpathy, A., Khosla, A., Bernstein, M., et~al. (2015).
\newblock Imagenet large scale visual recognition challenge.
\newblock {\em International journal of computer vision}, 115(3):211--252.

\bibitem[Salman et~al., 2020]{salman2020adversarially}
Salman, H., Ilyas, A., Engstrom, L., Kapoor, A., and Madry, A. (2020).
\newblock Do adversarially robust imagenet models transfer better?
\newblock {\em Advances in Neural Information Processing Systems}, 33.

\bibitem[Santurkar et~al., 2019]{santurkar2019image}
Santurkar, S., Ilyas, A., Tsipras, D., Engstrom, L., Tran, B., and Madry, A.
  (2019).
\newblock Image synthesis with a single (robust) classifier.
\newblock In Wallach, H., Larochelle, H., Beygelzimer, A., d'~Alch\'{e}-Buc,
  F., Fox, E., and Garnett, R., editors, {\em Advances in Neural Information
  Processing Systems}, volume~32, pages 1262--1273. Curran Associates, Inc.

\bibitem[Serre, 2019]{serre2019deep}
Serre, T. (2019).
\newblock Deep learning: the good, the bad, and the ugly.
\newblock {\em Annual review of vision science}, 5:399--426.

\bibitem[Szegedy et~al., 2015]{szegedy2015going}
Szegedy, C., Liu, W., Jia, Y., Sermanet, P., Reed, S., Anguelov, D., Erhan, D.,
  Vanhoucke, V., and Rabinovich, A. (2015).
\newblock Going deeper with convolutions.
\newblock In {\em Proceedings of the IEEE conference on computer vision and
  pattern recognition}, pages 1--9.

\bibitem[Szegedy et~al., 2014]{szegedy2013intriguing-properties-of-neural}
Szegedy, C., Zaremba, W., Sutskever, I., Bruna, J., Erhan, D., Goodfellow, I.,
  and Fergus, R. (2014).
\newblock Intriguing properties of neural networks.
\newblock In {\em International Conference on Learning Representations}.

\bibitem[Tsipras et~al., 2019]{tsipras2018robustness}
Tsipras, D., Santurkar, S., Engstrom, L., Turner, A., and Madry, A. (2019).
\newblock Robustness may be at odds with accuracy.
\newblock In {\em International Conference on Learning Representations}.

\bibitem[Valle-Perez et~al., 2019]{valle-perez2018deep}
Valle-Perez, G., Camargo, C.~Q., and Louis, A.~A. (2019).
\newblock Deep learning generalizes because the parameter-function map is
  biased towards simple functions.
\newblock In {\em International Conference on Learning Representations}.

\bibitem[Xie et~al., 2020]{xie2020adversarial}
Xie, C., Tan, M., Gong, B., Wang, J., Yuille, A.~L., and Le, Q.~V. (2020).
\newblock Adversarial examples improve image recognition.
\newblock In {\em Proceedings of the IEEE/CVF Conference on Computer Vision and
  Pattern Recognition}, pages 819--828.

\bibitem[Xie and Yuille, 2020]{xie2020intriguing}
Xie, C. and Yuille, A. (2020).
\newblock Intriguing properties of adversarial training at scale.
\newblock In {\em International Conference on Learning Representations}.

\bibitem[Yin et~al., 2019]{yin2019fourier}
Yin, D., Gontijo~Lopes, R., Shlens, J., Cubuk, E.~D., and Gilmer, J. (2019).
\newblock A fourier perspective on model robustness in computer vision.
\newblock {\em Advances in Neural Information Processing Systems},
  32:13276--13286.

\bibitem[Yosinski et~al., 2014]{yosinski2014transferable}
Yosinski, J., Clune, J., Bengio, Y., and Lipson, H. (2014).
\newblock How transferable are features in deep neural networks?
\newblock In {\em Advances in Neural Information Processing Systems}, pages
  3320--3328.

\bibitem[Zhang and Zhu, 2019]{zhang2019interpreting}
Zhang, T. and Zhu, Z. (2019).
\newblock Interpreting adversarially trained convolutional neural networks.
\newblock In {\em International Conference in Machine Learning}.

\end{thebibliography}


% Author biography without author photo.
% Author biography. Author biography. Author biography.
% Author biography. Author biography. Author biography.
% Author biography. Author biography. Author biography.
% Author biography. Author biography. Author biography.
% Author biography. Author biography. Author biography.
% Author biography. Author biography. Author biography.
% Author biography. Author biography. Author biography.
% Author biography. Author biography. Author biography.
% Author biography. Author biography. Author biography.

% \bio{figs/pic1}
% Author biography with author photo.
% Author biography. Author biography. Author biography.
% Author biography. Author biography. Author biography.
% Author biography. Author biography. Author biography.
% Author biography. Author biography. Author biography.
% Author biography. Author biography. Author biography.
% Author biography. Author biography. Author biography.
% Author biography. Author biography. Author biography.
% Author biography. Author biography. Author biography.
% Author biography. Author biography. Author biography.

% \bio{figs/pic1}
% Author biography with author photo.
% Author biography. Author biography. Author biography.
% Author biography. Author biography. Author biography.
% Author biography. Author biography. Author biography.
% Author biography. Author biography. Author biography.

\end{document}